\documentclass[lettersize,journal]{IEEEtran}
\usepackage{amsmath,amsfonts}
\usepackage{algorithmic}
\usepackage{algorithm}
\usepackage{array}
\usepackage{textcomp}
\usepackage{stfloats}
\usepackage{url}
\usepackage{balance}
\usepackage{verbatim}
\usepackage{cite}
\usepackage{adjustbox, booktabs, multirow, xcolor}
\usepackage{hyperref}
\usepackage{subcaption}
\usepackage{graphicx,authblk}
\usepackage[accsupp]{axessibility} 
\begin{document}

\title{Instance-level Few-shot Learning with Class Hierarchy Mining}

\author{Anh-Khoa Nguyen Vu, Thanh-Toan Do, Nhat-Duy Nguyen, Vinh-Tiep Nguyen*, Thanh Duc Ngo, Tam V. Nguyen
\thanks{Anh-Khoa Nguyen Vu, Nhat-Duy Nguyen, Vinh-Tiep Nguyen, Thanh Duc Ngo are with University of Information Technology, Ho Chi Minh City, Vietnam and Vietnam National University, Ho Chi Minh City, Vietnam (e-mail: khoanva@uit.edu.vn, duynn@uit.edu.vn, tiepnv@uit.edu.vn, thanhnd@uit.edu.vn)}

\thanks{Thanh-Toan Do is with Monash University, Clayton, VIC 3800, Australia (e-mail: toan.do@monash.edu)}%
\thanks{Tam V. Nguyen is with University of Dayton, Dayton, OH 45469, United States (e-mail: tamnguyen@udayton.edu)}%
\thanks{*Corresponding author}%
}

\markboth{IEEE TRANSACTIONS ON IMAGE PROCESSING}%
{Shell \MakeLowercase{\textit{et al.}}: A Sample Article Using IEEEtran.cls for IEEE Journals}


\maketitle

\begin{abstract}     
   Few-shot learning is proposed to tackle the problem of scarce training data in novel classes. However, prior works in instance-level few-shot learning have paid less attention to effectively utilizing the relationship between categories. In this paper, we exploit the hierarchical information to leverage discriminative and relevant features of base classes to effectively classify novel objects. These features are extracted from abundant data of base classes, which could be utilized to reasonably describe classes with scarce data. Specifically, we propose a novel \textit{superclass} approach that automatically creates a hierarchy considering base and novel classes as fine-grained classes for few-shot instance segmentation (FSIS). Based on the hierarchical information, we design a novel framework called Soft Multiple Superclass (\textit{SMS}) to extract relevant features or characteristics of classes in the same superclass. A new class assigned to the superclass is easier to classify by leveraging these relevant features. Besides, in order to effectively train the hierarchy-based-detector in FSIS, we apply the \textit{label refinement} to further describe the associations between fine-grained classes. The extensive experiments demonstrate the effectiveness of our method on FSIS benchmarks. Code is available online \footnote{\href{https://github.com/nvakhoa/superclass-FSIS}{github.com/nvakhoa/superclass-FSIS}}.
\end{abstract}

\section{Introduction}

A recently emerging approach, few-shot learning, is widely applied to address the problem of scarce data when training deep learning models. Specifically, few-shot learning is applied to resolve several vision tasks, from image-level tasks (classification~\cite{ben2021semantic,mall2021field,fei2021z},  
semantic segmentation~\cite{tian2020prior,lu2021simpler,Cheng_2021_ICCV,baek2021exploiting}) to instance-level tasks (object detection~\cite{tfa,max-margin, zhong2022pica,shaban2022few,lee2022few,rpn-attention}, instance segmentation~\cite{meta-rcnn,imtfa}). Moreover, this technique has also been developed in adaptation~\cite{lengyel2021zero,jhoo2021collaborative,zhang2021meta}, generative models~\cite{hinz2022charactergan,sheynin2021hierarchical,cheraghian2021synthesized, gu2021lofgan}, and action recognition~\cite{hong2021video,cao2020few,kumar2019protogan,mishra2018generative,patravali2021unsupervised,chen2021elaborative,memmesheimer2022skeleton}. Few-shot learning methods usually utilize two separate training sets; a base set contains numerous training data, and a novel one contains a limited number of samples per novel class. Therefore, few-shot models commonly have a training procedure with two stages, namely, base training and novel fine-tuning. Firstly, abundant training data of base classes is fed to a network for training to construct a feature space of base representatives. The network is then fine-tuned with a few samples of novel classes so that novel objects can be embedded within the feature space. This paper focuses on a practically desired, but rarely explored area, i.e., few-shot instance segmentation (FSIS).

\begin{figure}[!t]
    \centering
    \includegraphics[width=1\linewidth]{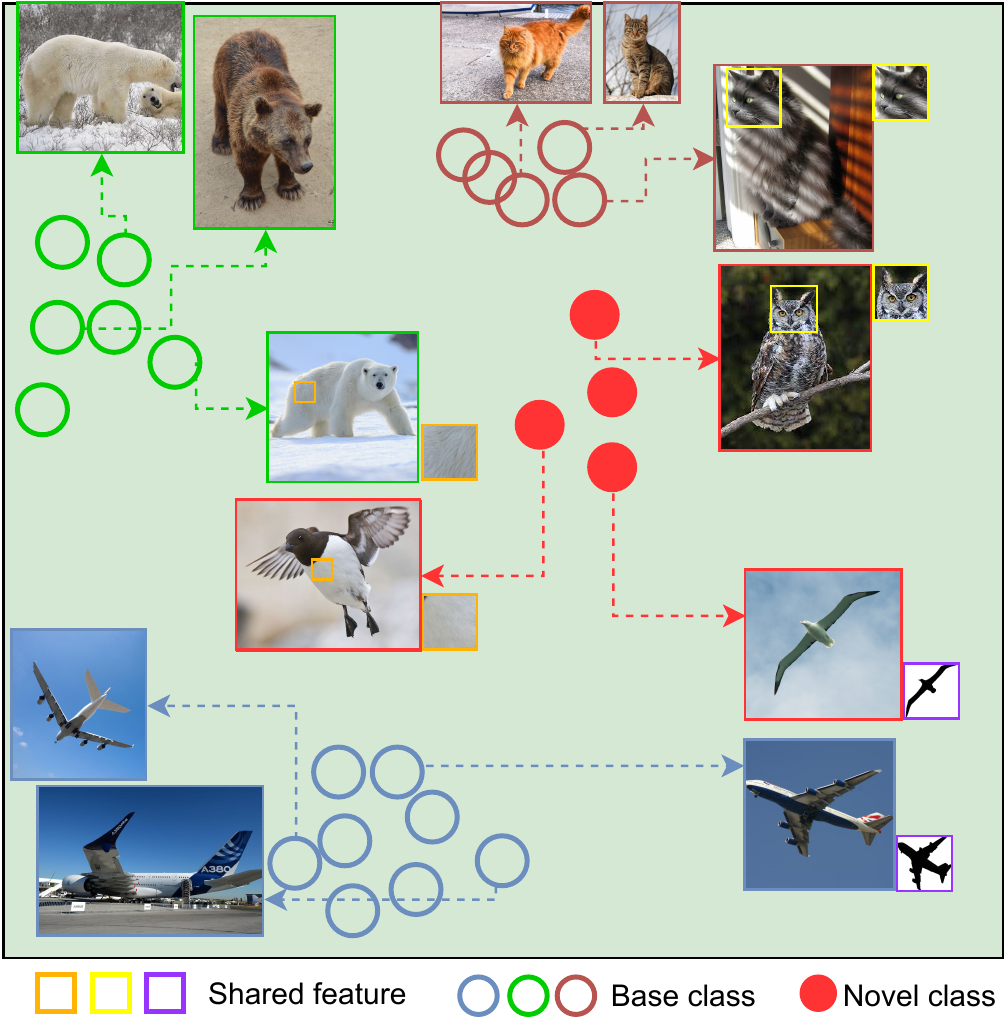}
    \caption{An exemplary illustration of our proposal. For instance, there are 3 base classes (`bear’, `cat’, and `airplane’) and 1 novel classes (`bird’). Here, the novel `bird’ class is represented in the feature space of 3 base classes including `airplane’, `cat’, and `bear’ by leveraging local features (`face', `fur') and global features (`shape').}
    \label{fig:teaser}
\end{figure}

Pioneering works \cite{meta-rcnn,yolo-reweighting,rpn-attention,hu2021dense} focus on the attention mechanisms for identifying new objects. Another branch develops data augmentation \cite{trans-int,max-margin,khandelwal2021unit} to alleviate the scarcity of novel classes. However, these current methods focus little on exploiting relevant information between base and novel classes that do not need to rely remarkably on data. Extracting associative information from well-learned base and novel classes will effectively support the model to make good predictions in the novel domain.

In contrast to computers, humans can quickly learn new objects with little receptive information about them by combining features of known things to form the features of a new object. For example, if we conceptually  have prior knowledge of trains and cars, we can logically deduce buses from some exemplary pictures. It means that trains and cars resemble buses in part on external designs. Therefore, we can take advantage of relevant information of known objects to recognize a new object. 

A recent work for object detection \cite{zhu2021semantic} also has an idea for mining information on semantic relations between categories. Still, they used a large amount of additional text data to learn associations.

In this work, we present a new approach to exploit the relations between categories without adding or augmenting data. The method, called the \textbf{superclass} approach, groups base and novel classes with highly similar characteristics  into one superclass. For example, in Fig. \ref{fig:teaser}, the base classes (‘bear', ‘airplane', and ‘cat') and the novel class (‘bird') are assigned to one superclass. There are high similarity features between these classes such as local features of animals (‘fur’, ‘face’) and global features of flying objects (‘shape’).

The detector can productively classify novel classes in the superclass via leveraging such shared features which are well-trained on base classes with abundant data. In other words, the model can utilize this rich diversity of features from both base and novel classes instead of using features from only novel classes with limited annotation data. Specifically, we propose a novel framework (named \textbf{Soft Multiple Superclass – SMS}) for FSIS which is embedded in a hierarchical structure with the aim of learning relevant features. Moreover, to effectively train the superclass model, we introduce \textbf{Refinement Label (RL)} method to better describe associations between fine-grained classes. Finally, our approach (SMS + RL) outperforms the current state-of-the-art in FSIS. We also provide extensive ablation studies for our methods. Our contributions can be summarized as follows:

\begin{itemize}
    \item We propose a new approach to efficiently exploit hierarchy features for FSIS, called the superclass approach.

    \item We present an end-to-end deep network utilizing the hierarchical information to effectively detect novel objects by leveraging shared features between base and novel classes.
    
    \item We introduce a label refinement method to enhance the association between anchor base and novel classes in a superclass.
    
\end{itemize}

\section{Related Work}
\label{sec:related}

\subsection{Few-shot semantic segmentation}
Methods \cite{panet,Min_2021_ICCV,tian2020prior,wu2021learning,lu2021simpler,Cheng_2021_ICCV,baek2021exploiting} are recently proposed to deal with semantic segmentation when the training data is limited. PANet \cite{panet} focuses on exploring prototype representations from support images. The method predicts segmentation masks via non-parametric metric learning and forces the model to learn a consistent embedding space for queries and support prototypes via the prototype alignment regularization. Min \textit{et al.}~\cite{Min_2021_ICCV} propose HSNet which uses 4D convolutions to exploit multi-level feature correlation. Wu \textit{et al.}~\cite{wu2021learning} and Xie \textit{et al.}~\cite{xie2021few} design memory networks to alleviate the problem. Yang \textit{et al.}~\cite{Yang_2021_ICCV} create a pseudo mask label and apply it to the join training stage. In this way, they exploit the relevant information between classes via the loss function. Different from this work, our method uses the information to construct hierarchical information and applies it to the network architecture during both inference and training.

\subsection{Few-shot object detection and instance segmentation}
Different approaches have been proposed for few-shot object detection (FSOD) ~\cite{zhong2022pica,shaban2022few,lee2022few,rpn-attention,yolo-reweighting,zhu2021semantic} and instance segmentation (FSIS)~\cite{meta-rcnn,imtfa,fgn} and they can be roughly divided into different categories as follows:

\noindent\textbf{Attention mechanism.} Early works \cite{meta-rcnn,yolo-reweighting} guide to encourage detectors to pay more attention to objects of interest via attention information gained by a meta-learner. The difference is that Meta R-CNN \cite{meta-rcnn} works on an objective with two tasks few-shot object detection and segmentation. Fan \textit{et al.} \cite{rpn-attention} propose an attention module called Attention-RPN, Multi-Relation Detectors combined with massive FSOD data to match query proposals and support objects. DCNet \cite{hu2021dense} uses the Dense Relation Distillation module and Context-aware Feature Aggregation to aggregate information from local and global features via the attention mechanism. Lang \textit{et al.} \cite{lang2022learning} introduce the BAM method to alleviate the bias of seen classes over new coming classes by adding a base-learner branch to the conventional meta-learner model with an ensemble function. {Based on a transformer framework, RefT~\cite{han2023reference} recently introduces two attention mechanisms that are
applied at feature and instance levels to explore the relationship between support and query features. }

\noindent\textbf{Data augmentation.} Transformation Invariant Principle (TIP) \cite{trans-int} is developed to train two networks, Guidance Extractor and Detector,  to improve the generalization ability via learning invariants generated by a transformation augmentation. Different from TIP, Li \textit{et al.} \cite{max-margin} present a class margin equilibrium to optimize feature space partition and fine-tune this margin equilibrium with online data augmentation based on feature disturbance. UniT \cite{khandelwal2021unit} proposes an approach based on using additional image-level data for novel classes. This semi-supervised method has yielded remarkable results for FSOD.  However, the methods in this approach do not match our objective for making a comparison in this work. 

\noindent\textbf{Model-based approach.} TFA \cite{tfa} simply uses Faster RCNN \cite{faster-rcnn} with two-stage fine-tuning which has superiority in performance over previous works. In FSCE \cite{sun2021fsce}, instead of fine-tuning the last two layers, FSCE adapts Faster RCNN with the contrastive proposal encoding to exploit inter-class relationships for learning more robust object feature representations from fewer samples. Nguyen \cite{nguyen2022ifs} extends Mask-RCNN to propose iFS-RCNN which improves the fine-tuning stage for the new classes. The model leverages the Bayesian function and a novel uncertainty-guided bounding predictor to specify novel objects. Siamese Mask R-CNN \cite{oneshot-instance} proposes an architecture for few-shot instance segmentation via a siamese network backbone and feature matching which measures the similarity between the reference and query image. In \cite{fgn}, Zhibo \textit{et al.} design Fully Guided Network (FGN) which has three sub-guidance modules in order to effectively use the signals from the support set and adapt better to the novel domain. Based on TFA, Ganea \textit{et al.} present MTFA \cite{imtfa} that leverages a two-stage fine-tuning approach and adds a mask prediction branch as Mask RCNN extends Faster RCNN. They also propose iMTFA which replaces the fixed feature extractor with an instance feature extractor and averages all per-class instances to create a new embedding. Hence, they can train in novel classes without providing instance masks. {Recently, DTN~\cite{ wang2022dynamic} introduces new transformer architecture to directly segment instances in images. Moreover, they also reduce noise by using cross-attention in Semantic-induced Transformer Decoder. Note that DTN~\cite{ wang2022dynamic} uses the other setting in COCO data, which is not commonly used in previous FSIS papers.}

\noindent\textbf{Relation information.} In the context of research about the relation between base and novel classes. Zhu \textit{et al.} exploit the relation of classes in \cite{zhu2021semantic} via extra data. Specifically, they propose SRR-FSD that uses external text information to learn the relation between base and novel classes. Features that carry relation information are augmented and support the classifier in recognizing the objects.  When it comes to leveraging high-level semantic relationships among objects, Yanan \textit{et al.} \cite{li2020context} propose a framework leveraging the idea of the superclass for zero-shot detection by using context information via the multiplication of two probabilities of superclass prediction output and label prediction. Ashima \textit{et al.} \cite{garg2022hiermatch} exploit the label hierarchy by building different predefined hierarchical levels and using coarse label hierarchies for semi-supervised learning. Differently, instead of using extra data or leveraging context information surrounding objects in the image or predefined hierarchical labels, we leverage the idea of the superclass by extracting relevant features within categories in the same superclass to constitute hierarchical labels for ease of classifying new objects without further doing nothing. In the next section, we introduce our proposed framework and superclass idea.

\section{Superclass Framework}
\label{sec:framework}

\begin{table}[!t]
\centering
\small
\begin{tabular}{|c|l|}
\bottomrule
Notation & Meaning \\\hline
  $\mathbb{X}_c$ & The set instances of class $c$\\
  $K$ & Number of shots \\
  $C_b$ & Base classes \\
  $C_n$ & Novel classes \\
  $N_s$ & Number of superclasses \\
  $N_{g_i}$ & Number of fine-grained classes in the $i^{th}$ superclass \\\hline
  $f(\cdot)$ & The feature extractor  \\
  $h(\cdot)$ & The classifier in the base model   \\
  $S(\cdot)$ & The superclass classifier   \\
  $G_i(\cdot)$ & The fine-grained classifier is for the $i^{th}$ superclass \\\hline
  $\gamma$ & The hyperparameter to create the superclasses \\
  $F$ & The extracted features from $f(\cdot)$\\
  $u_c$ & The representative of the class $c$ \\\hline
  \multirow{2}{*}{$\hat{\mathcal{T}}$} & $\hat{\mathcal{T}}= h(F)$ is logit output before \\ & the softmax layer\\\hline
   \multirow{3}{*}{$s$}  &  $s \in \{0, 1\}^{N_s}$ is multi-hot encoding where;\\ & $s_i$ = 1 if an instance belongs to \\ & the $i^{th}$ superclass, 0 otherwise \\\hline

  \multirow{3}{*}{$g_{i}$ }  & $g_i \in \{0, 1\}^{N_{g_i}}$ is one-hot vector where; \\ & $g_{ij}$ = 1  if an object belongs to $j^{th}$ fine-grained \\& class in $i^{th}$ superclass, 0 otherwise
  \\\toprule
\end{tabular}
\caption{Notations and their corresponding meanings.}
\label{tab:notations}
\end{table}
\subsection{Problem definition}
We summarize notations in Table~\ref{tab:notations}. In FSIS, training data composes of base classes $C_b$ having abundant instances and novel classes $C_n$ with few instances. Few-shot approaches explore the training data and simultaneously segment objects from both base and novel classes. Our framework is based on Mask RCNN \cite{he2017mask}. For the first training, the model is optimized through the loss function:

\begin{equation}
\label{eq:base_loss}
    L_\text{total} = L_\text{cls} + L_\text{bbx} + L_\text{rpn} + L_\text{mask}
\end{equation}
where $L_\text{cls}$ , $L_\text{bbx}$, $L_\text{rpn}$,  $L_\text{mask}$ denote the classification, bounding box regression, region proposal and segmentation loss, respectively. The cross-entropy loss for classification of each instance in Equation \ref{eq:base_loss} is defined:
\begin{equation}
\label{eq:base_cls_loss}
   L_\text{cls} =  -  \sum_i^{|C_b|} q_i\log(\hat{q}_i),
\end{equation}
where $q$ is the groundtruth label vector 
and $\hat{q}_i$ is the probability of the $i$-th class in $C_b$.  In the next section, we present the proposed idea and how to apply it in the network architecture for  the fine-tuning stage. 

\subsection{Superclass Mining}

\textbf{Superclass hierarchy.} 
The core idea is to group categories in $C_b$ and $C_n$ into $N_s$ superclasses and form them into a hierarchical structure. 
The identified categories in a group contain highly relevant information. Therefore, few-shot approaches based on hierarchical structures can leverage features of associated base classes to represent novel ones.

\begin{figure}[!t]
    \centering
    \includegraphics[width=1\linewidth]{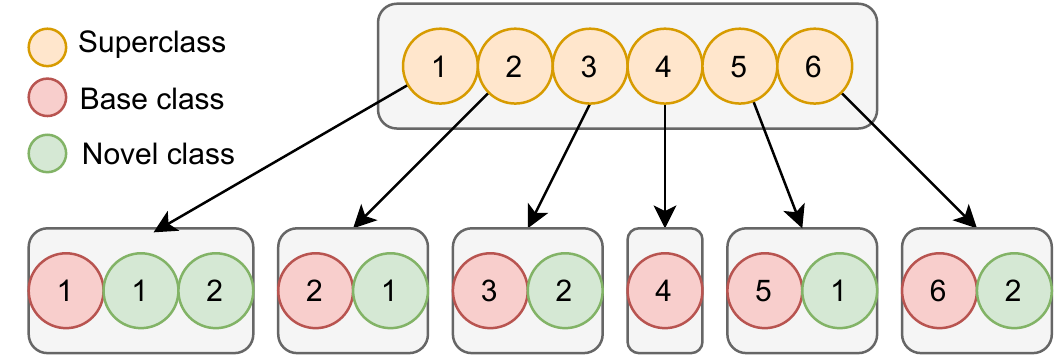}
    \caption{An exemplary hierarchy of the superclass approach. This hierarchy has 6 base classes and 2 novel classes with $\gamma$ = 3.}
    \label{fig:hierarchy}
\end{figure}
\noindent\textbf{Similarity measurement.} To measure the similarity between base and novel classes, we use a pretrained network in first training, which is optimized by Equation \ref{eq:base_loss}, to  predict novel instances. The training data of novel class is not used when training this network.

Let ${\hat{\mathcal{T}}} = h(f(\mathbb{X}_c))$ is the logit output, $c \in C_n$, and $|\mathbb{X}_c| = K $. We calculate the class representative $u_c = \frac{1}{K} \sum_{k=1}^K {\hat{\mathcal{T}}}_k$ for each novel class $c$. Then, we choose the top $\gamma$ base classes that are similar to $u_c$. The novel class $c$ is assigned to $\gamma$ supper classes corresponding to $\gamma$ selected base classes. Accordingly, a superclass $C_{s}$ is created by grouping only a single base class (base anchor) and several novel classes. All classes within a superclass are then considered as fine-grained classes. 

Following the above approach, we consider a base representative as an anchor containing informative features to describe a new class. In other words, a novel class therefore is interpreted by $\gamma$ base classes via prior knowledge of the pretrain model. Fig.  \ref{fig:hierarchy} illustrates a hierarchical tree for the superclass approach in which each novel class is interpreted by three base classes ($\gamma$ = 3).

To implement the superclass idea, we can train two separate networks. However, this practice is costly due to the multiple training procedure. Therefore, we create a framework that leverages the idea of the superclass to train an end-to-end model. The network in the first training is modified to use the related information between classes for the fine-tuning stage. 

To be specific, the classifier $h(\cdot)$ is now replaced by a superclass classifier $S(\cdot)$ and $N_s$ fine-grained classifiers; the fine-grained classifier $G_i(\cdot)$ is for the $i^{th}$ superclass. 

Let $\hat{s} = \textit{sigmoid}({S(f(\mathbb{X}_c))})$ be the probability output vector of  $S(\cdot)$; $c$ can be either a base or novel class. The loss function for the superclass classifier $S(\cdot)$ is defined as 
\begin{equation}
\label{eq:super_loss}
  L_{s} =  - \sum_i^{N_s} s_i\log(\hat{s_i}) + (1-s_i)\log(1-\hat{s_i})
\end{equation}

\begin{figure}[!t]
    \centering
    \includegraphics[width=1\linewidth]{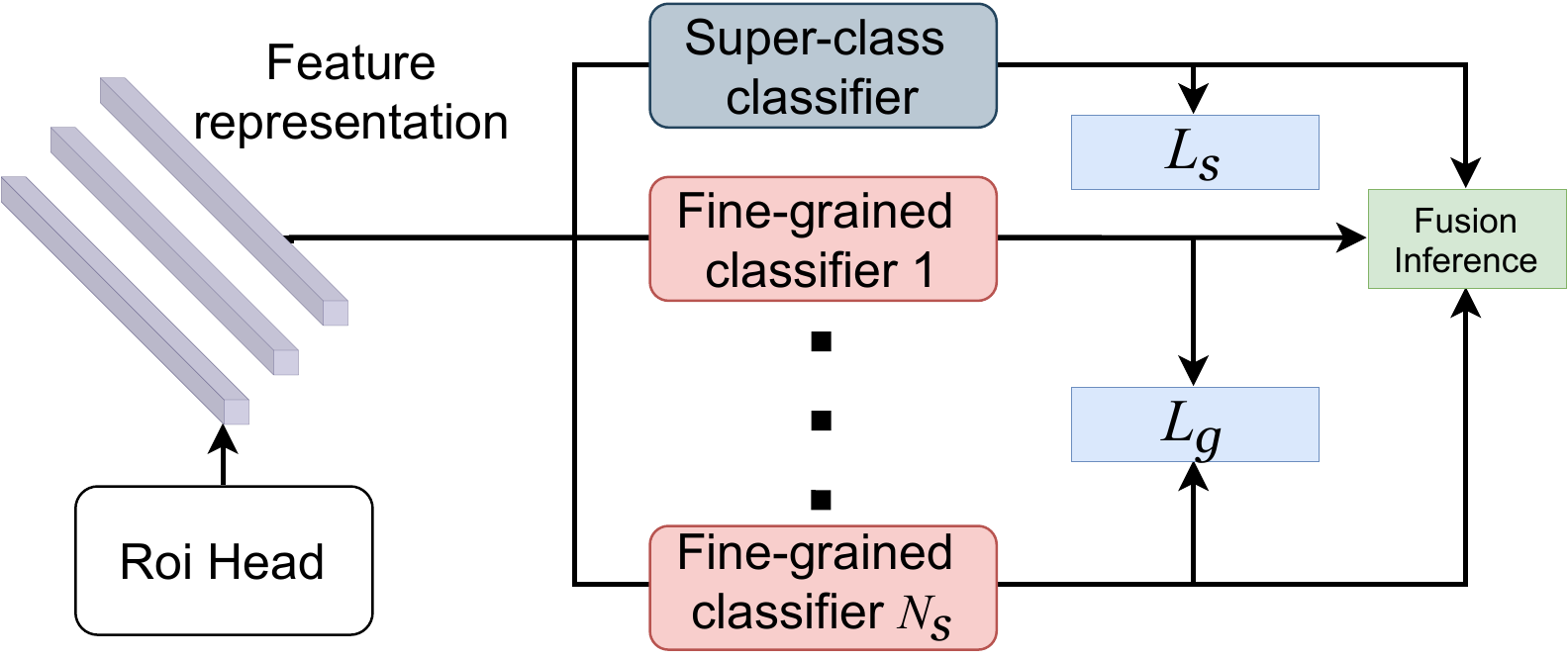}
    \caption{The proposed superclass framework. The feature maps from Roi Head are simultaneously fed into the superclass classifier and the fine-grained classifiers to learn the corresponding task via $L_s$ and $L_g$, respectively. The final decision of the framework is computed by fusing outputs from the superclass classifier and fine-grained classifiers.}
    \label{fig:framework}
\end{figure}

Let $\hat{g}_i = \textit{softmax}(G_i(f(\mathbb{X}_c))) $ be the probability output vector of $G_i(\cdot)$. The loss function of all fine-grained classifer is defined as
\begin{equation}
\label{eq:fine-grained_loss}
 L_{g} = -\sum_i^{N_s} \sum_j^{N_{g_i}} g_{ij}\log(\hat{g}_{ij}),
\end{equation}

The final loss function for the superclass model is:
\begin{equation}
\label{eq:novel_loss}
   L_\text{total} =  \alpha L_{s} + \beta L_{g} + L_\text{bbx} + L_\text{rpn} + L_\text{mask}.  
\end{equation}

The weights $\alpha$ and $\beta$ are added to balance the superclass loss function $L_s$ and the fine-grained class loss function $L_g$ with others during training.

\noindent\textbf{Fusion inference.} In the testing, we gather the output from the superclassifier and the fine-grained classifiers to make the final prediction. Specifically, the output score ${p}_c$ of the class $c$ in dataset is synthesized by the predictions of fine-grained classes and respective superclass and is defined as:
\begin{equation}
\label{eq:predict_func}
\begin{aligned}
    p_c &= \sum_i^{N_s} \mathcal{Q}_{ic} * \hat{s}_i \\
\end{aligned}
\end{equation}

Where ${Q}_{ic}$ is the predicted probability for the class $c$ by the fine-grained classifier $G_i$ if the $i^{th}$ superclass contains the class $c$. Otherwise ${Q}_{ic}$ = 0. $\hat{s}$ is an output vector of the superclass predictor which is presented in the next section.

From the above definitions, we propose \textbf{Soft Multiple Superclass (SMS)} which is a general framework to train an end-to-end superclass model and fuse the predictions from classifiers. This framework uses $\gamma > 1$ to create the hierarchy. The overview superclass framework is shown in Fig.  \ref{fig:framework}. However, we also consider two stricter versions when $\gamma = 1$. In the next section, we show our strict frameworks for the superclass approach and their drawbacks when compared to SMS. Equation \ref{eq:super_loss} can be updated depending on different variants in our framework to suit the model target.

\subsection{Superclass Variants}
\label{subsec:Framework Variants}

\textbf{Hard Single Superclass (HSS)} is our first version in which a novel class is assigned to only a superclass, i.e., $\gamma=1$. The inference output of the model is strongly based on $S(\cdot)$. 

As a result, Equation \ref{eq:super_loss} in this version is redefined as follows, respectively:

\begin{equation}
\label{eq:super_loss_hard_version}
 L_{s} = -\sum_i^{N_s} s'_{i}\log(\hat{s}'_i) ),
\end{equation}

\noindent where $s'$ and $\hat{s}'$ = \textit{softmax}($S(F)$) are the one-hot vector label and the prediction  of the $i^{th}$ superclass, respectively. And $\hat{s}$ in Equation \ref{eq:predict_func} equals $\textit{onehot}(\underset{}{\operatorname{argmax}}(\hat{s}'), N_s)$. The function $\textit{onehot}(\textit{index}, N)$ is utilized to generate a one-hot $N$-dimensional vector with value of 1 at \textit{index}.

\noindent\textbf{Soft Single Superclass (SSS)} is an upgrade of HSS. We realize that the final inferences of our model are only based on absolute decisions of the superclass classifier like the HSS framework. This leads the 
model to yield incorrect predictions about novel objects in an image. In this version, we use Equation \ref{eq:super_loss_hard_version} of \textit{HSS} to train the superclass classifer and the output probability $\hat{s}$ in Equation \ref{eq:predict_func} equals $\textit{sigmoid}(S(f(\mathbb{X}_c)))$.

\noindent\textbf{Drawbacks.} Both above versions are based on only one base class ($\gamma=1$) to describe the relation between the base anchors and fine-grain classes. However, if the information is wrong for any reason, it will reduce the generalization ability of the model over the novel domain. To this a question about our framework is probably raised:  Would the model learn better if we use the characteristics of different base classes to describe a novel class?  From the empirical study, we have observed that our model predicts better when using multiple superclasses (e.g., $\gamma = 2 $) to describe a new class than a single superclass. Hence, our final framework is \textbf{Soft Multiple Superclass (SMS)}.

\begin{table*}[]
\begin{subtable}{0.5\linewidth}
\centering
\begin{tabular}{c|cc|cc}
\bottomrule
\multirow{2}{*}{Method} & \multicolumn{2}{c|}{Segmentation}                  & \multicolumn{2}{c}{Detection}                     \\
                        & \multicolumn{1}{c}{AP} & \multicolumn{1}{c|}{AP50} & \multicolumn{1}{c}{AP} & \multicolumn{1}{c}{AP50} \\\hline
HSS                     & 3.82                   & 7.80                     & 3.30                   & 7.63                     \\
SSS                     & 7.23                   & 11.95                    & 6.52                   & 11.85                    \\
SMS ($\gamma=2$)        & \textbf{10.58}                  & \textbf{19.16}                    & \textbf{10.20}                  & \textbf{19.71}                    \\
SMS ($\gamma=3$)        & 10.01                  & 18.49                    & 9.78                   & 19.46                    \\
SMS ($\gamma=5$)        & 8.22                   & 15.74                    & 8.00                   & 16.75     \\\toprule              
\end{tabular}
\caption{}
\label{table:ablations_framework_variants}
\end{subtable} 
\begin{subtable}{0.5\linewidth}
\centering
\begin{tabular}{c|cc|cc}
\bottomrule
\multirow{2}{*}{$\gamma$} & \multicolumn{2}{c|}{Segmentation} & \multicolumn{2}{c}{Detection} \\
                        & AP            & AP{50}          & AP              & AP{50}           \\\hline
2                    & 10.84           & 18.91          & 10.58         & 19.36         \\
3                    & 11.28           & 19.96          & 11.15         & 20.81         \\
5                    & 11.80           & 20.80          & 11.71         & 21.64         \\
10                   & 11.70           & 20.87          & 11.60         & 21.64         \\
15                   & \textbf{11.80}           & \textbf{21.06}          & \textbf{11.74}         & \textbf{21.86}         \\  
\toprule
\end{tabular}
\caption{}
\label{tab:Ablations_gamma}
\end{subtable}
\caption{Ablations variants of our method and base anchors on 10-shot of COCO dataset. \textbf{(a)} shows comparison between different variants of the superclass approach and the effectiveness of increasing $\gamma$ \textbf{without Label Refinement}. \textbf{(b)} shows the performance of the SMS variant \textbf{with Label Refinement} is improved as $\gamma$ is grown and peaks at $\gamma$ = 15. Numbers in bold indicate the best performance.}
\end{table*}

\begin{figure}[!t]
    \centering
    \includegraphics[width=0.7\linewidth]{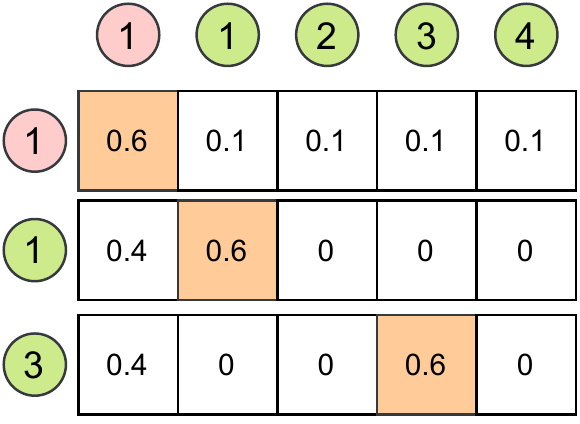}
    \caption{Illustration of Label Refinement. Here we illustrate a superclass with five classes (1, 1, 2, 3, 4) including the base class (pink circle) and the four novel classes (green circles), and three instances corresponding to three rows (one base and two novel class instances). Rows are distribution ground-truth for instances with $\epsilon = 0.4$. As illustrated, an instance belonging to the base anchor ($1^{st}$ row) uses Equation \ref{eq:smoothing}  to refine the label. Equation \ref{eq:refinement} is applied to objects in the remaining fine-grained classes ($2^{nd}$ and $3^{rd}$ rows).}
    \label{fig:LR_example}
\end{figure}
\subsection{Label Refinement for SMS Variant}
\label{subsec:LR}

Though we have formulated the superclass idea to the model, the well-learned representations can dominate the new representations when optimizing the model. This situation can happen since the annotation is labeled in a ``hard" description and not good in describing all relevant information between the base anchor and the fine-grained classes in a superclass.  For example, our experimental evidence in Table \ref{table:ablations_framework_variants} shows that the performance drops as we increase base anchors $\gamma$ up to $3$ or $5$.

Assume that in the $i^{th}$ superclass, there is an instance $x_z$ of the $z^{th}$ fine-grained class and the base anchor is the $d^{th}$ fine-grained class. We require $z \neq d$. Let $o_i = G_i(f(x_z))$ is the logit vector output before the softmax layer. The groundtruth label $g_{ij} $ is created by Dirac delta function, which equals 1 if $j = z$, 0 otherwise. When the model is optimized by using the cross-entropy loss function with this label, it leads to the logit $o_{ij} \gg o_{iz}$ if $j \neq z$ and requires the model to learn discriminative information between fine-grained classes instead of looking for relevant features that support novel representatives. Basically, we do not expect this situation to happen because the fine-grained classes within a superclass must leverage information of each other to make the predictions.

To tackle this issue, we propose a mechanism called \textbf{Label Refinement (LR)} for encouraging the model to learn similar representations between the base anchor and fine-granied classes in a superclass. In detail, we adapt label smoothing \cite{szegedy2016rethinking} to refine the annotation for fine-grained categories. The origin label smoothing is defined as:

\begin{equation}
\label{eq:smoothing}
    g_{ij} = \begin{cases}
      1 - \epsilon, \text{if }  j = z \\
\frac{\epsilon}{N_{g_i} - 1}, \text{otherwise}
    \end{cases}.
\end{equation}

\begin{table*}[]
\centering
\adjustbox{max width=1.0\textwidth}{
\begin{tabular}{c|l|cccccc|cccccc}
\bottomrule
\multirow{2}{*}{Shots} & \multirow{2}{*}{Method} & \multicolumn{6}{c}{Segmentation}            & \multicolumn{6}{|c}{Detection}                \\\cline{3-14}
                      &                             & AP    & AP50  & AP75  & APs  & APm   & APl   & AP    & AP50  & AP75  & APs  & APm   & APl   \\\hline
\multirow{4}{*}{1}     & iMTFA~\cite{imtfa}                          & 2.81  & 4.72  & 2.90  & 0.85 & 2.37  & 5.01  & 3.23  & 5.89  & 3.13  & 1.24 & 2.89  & 5.13  \\
                      & MTFA~\cite{imtfa}                              & 2.34  & 3.99  & 2.46  & 0.69 & 2.20  & 3.89  & 2.10  & 4.07  & 1.99  & 0.96 & 2.16  & 3.16  \\
                      & iFS-RCNN~\cite{nguyen2022ifs}                              & 3.95  & \_  & \_  & \_ & \_  & \_  & 4.54  & \_  & \_  & \_ & \_  & \_  \\
                      & RefT~\cite{han2023reference} & \underline{5.10} & \_ & \_ & \_ & \_ & \_ & \underline{5.20} & \_ & \_ & \_ & \_ & \_ \\
                      & MTFA*(baseline)             & 3.28  & 5.28  & 3.50  & 1.39 & 2.19 & 5.42  & 3.33  & 5.71  & 3.50  & 1.74 & 2.53  & 4.66  \\
                      & SMS+LR (Ours)                   & \textbf{5.43}  & \textbf{8.72}  & \textbf{5.92}  & \textbf{1.62} & \textbf{4.41}  & \textbf{8.91}  & \textbf{5.29}  & \textbf{9.58}  & \textbf{5.61}  & \textbf{2.22} & \textbf{4.59}  & \textbf{7.74}  \\\hline
\multirow{6}{*}{5}     & MRCN+ft-full~\cite{meta-rcnn}           & 1.30  & 2.70  & 1.10  & 0.30 & 0.60  & 2.20  & 1.30  & 3.00  & 1.10  & 0.30 & 1.10  & 2.40  \\
& Meta RCNN~\cite{meta-rcnn} & 2.80 & 6.90 & 1.70 & 0.30 & 2.30 & 4.70  & 3.50  & 9.90  & 1.20  & 1.20 & 3.90  & 5.80  \\
                      & iMTFA~\cite{imtfa}   & 5.19  & 8.65  & 5.36  & 1.58 & 4.41  & 9.52  & 6.07  & 11.15 & 5.86  & 2.35 & 5.19  & 9.82  \\
                      & MTFA~\cite{imtfa}                    & 6.38  & 11.14 & 6.46  & 1.92 & 5.20  & 10.47 & 6.22  & 11.63 & 6.01  & 2.77 & 5.57  & 9.54  \\
                      
                      & iFS-RCNN~\cite{nguyen2022ifs}                              & 8.80  & \_  & \_  & \_ & \_  & \_  &  \underline{9.91}  & \_  & \_  & \_ & \_  & \_  \\
                      & RefT~\cite{han2023reference} & \textbf{12.70} & \_ & \_ & \_ & \_ & \_ & \textbf{14.10} & \_ & \_ & \_ & \_ & \_ \\
                      & MTFA*(baseline)              & 8.31  & 14.01 & 8.72  & 3.03 & 6.29 & 13.89 & 8.46  & 14.35 & 8.88  & 4.33 & 7.21  & 12.53 \\
                      & SMS+LR (Ours)                  & \underline{9.32}  & \textbf{16.96} & \textbf{9.16}  & \textbf{2.85} & \textbf{7.05}  & \textbf{15.79} & {9.43}  & \textbf{18.03} & \textbf{8.99}  & \textbf{4.02} & \textbf{7.93}  & \textbf{14.70} \\\hline
\multirow{5}{*}{10}    & MRCN+ft-full~\cite{meta-rcnn}            & 1.90  & 4.70  & 1.30  & 0.20 & 1.40  & 3.20  & 2.50  & 5.70  & 1.90  & 2.00 & 2.70  & 3.90  \\ & Meta RCNN~\cite{meta-rcnn} & 4.40  & 10.60 & 3.30  & 0.50 & 3.60 & 7.20  & 5.60  & 14.20 & 3.00  & 2.00 & 6.60  & 8.80  \\
                      & iMTFA~\cite{imtfa}                    & 5.88  & 9.81  & 6.07  & 1.75 & 5.09  & 10.60 & 6.97  & 12.72 & 6.70  & 2.76 & 5.96  & 11.06 \\
                      & MTFA~\cite{imtfa}                    & 8.36  & 14.58 & 8.46  & 2.45 & 6.62  & 13.36 & 8.28  & 15.25 & 8.14  & 3.67 & 7.34  & 12.46 \\
                      
                      & iFS-RCNN~\cite{nguyen2022ifs}                              & 10.06  & \_  & \_  & \_ & \_  & \_  &  \underline{12.55}  & \_  & \_  & \_ & \_  & \_  \\
                      
                      & RefT~\cite{han2023reference} & \textbf{17.50} & \_ & \_ & \_ & \_ & \_ & \textbf{18.90} & \_ & \_ & \_ & \_ & \_ \\
                      & MTFA* (baseline)                 & 9.79  & 17.39 & 9.98  & 3.33 & 8.23 & 17.83 & 9.91  & 17.76 & 9.62  & 4.99 & 9.03  & 16.05 \\
                      & SMS+LR (Ours) & \underline{12.19} & \textbf{21.64} & \textbf{12.34} & \textbf{3.58} & \textbf{9.53}  & \textbf{20.92} & {12.31} & \textbf{22.76} & \textbf{11.67} & \textbf{5.12} & \textbf{11.01} & \textbf{19.33} 
\\\toprule
\end{tabular}}
\caption{Few-shot instance segmentation performance on the COCO novel test set. `$\_$' indicates results are not reported. Our baseline IMTFA* is fine-tuned by using our backbone without class-agnostic, and its hyperparameters are tuned to optimize the few-shot instance segmentation task. The best and second best results are denoted in \textbf{bold} and \underline{underline}, respectively.}
\label{table:novel_set_coco}
\end{table*}

\begin{table}[!t ]
\centering
\begin{tabular}[width=\linewidth]{c|c|l}
\bottomrule
Shot                & Superclass        & Fine-grained classes                           \\\hline
\multirow{6}{*}{1}  & \multirow{2}{*}{1} & \textbf{truck}, bicycle, m.bike,                    \\
                    &                    & bus, train, boat,  car                         \\\cline{2-3}
                    & 2                  & \textbf{fire hydrant}, car, m.bike, bottle          \\\cline{2-3}
                    & 3                  & \textbf{bear}, cat, dog, sheep, cow                     \\\cline{2-3}
                    & \multirow{2}{*}{4}        & \textbf{elephant}, horse, sheep, cow, \\
                      &              & person, bird, cat \\\hline
\multirow{5}{*}{5}  & \multirow{2}{*}{1} & \textbf{truck}, bicycle, m.bike,                     \\
                    &                    & bus,   train, boat,  car                      \\\cline{2-3}
                    & \multirow{2}{*}{2}                  & \textbf{fire   hydrant}, car, m.bike, \\
                    & & potted plant  \\\cline{2-3}
                    & 3                  & \textbf{bear}, dog, sheep, cow                          \\\cline{2-3}
                    & 4                  & \textbf{elephant}, horse, sheep, cow                    \\\hline
\multirow{5}{*}{10} & \multirow{2}{*}{1} & \textbf{truck}, bicycle, m.bike,                     \\
                    &                    & bus, train, boat, car                         \\\cline{2-3}
                    & 2                  & \textbf{fire hydrant}, m.bike, potted plant         \\\cline{2-3}
                    & 3                  & \textbf{bear}, dog, sheep, cow                          \\\cline{2-3}
                    & 4                  & \textbf{elephant}, horse, sheep, cow  
                    \\\toprule
\end{tabular}
\caption{Superclass information. The fine-grained classes in 4 superclasses with  $\gamma$ = 3. Here, bold texts are the base anchors with a supperclass such as`truck', `fire hydrant', `bear', `elephant'.}
\label{tab:fg_classes}
\end{table}

The $\epsilon$ scalar is added to prevent the model from overfitting and $N_{g_i}$ is the number of fine-grained classes in the $i^{th}$ superclass. In addition, $\epsilon$ also describes the similarity between the base anchor and fine-grained classes. By following Equation \ref{eq:smoothing}, we can create related features between fine-grained classes in the same group. However, the novel representatives are not well learnt to support each other and can make the model give erroneous predictions. Therefore, we leverage and re-define Equation \ref{eq:smoothing} to describe the relevant information from fined-grained classes to the base anchor as follows:
\begin{equation}
    g_{ij} = \begin{cases}
      1 - \epsilon, \text{if } j = z\\
      \epsilon, \text{if } j = d \\
      0, \text{otherwise}
    \end{cases}
\label{eq:refinement}
\end{equation}
, where the distribution groundtruth for the $i^{th}$ superclass has the $\epsilon$ to indicate the linking between the base anchor and the other fine-grained classes. 
The method creates the groundtruth distribution for a base anchor and other fine-grained classes by using using Equations \ref{eq:smoothing} and \ref{eq:refinement}, respectively (shown in Fig.  \ref{fig:LR_example}). To integrate the method for the SMS variant, we add another cross-entropy loss function for Label Refinement.

\section{Experiments}
\label{sec:exp}
In this section, we provide evidences to demonstrate the efficiency of our superclass model for dealing with the few-shot instance segmentation. We start from data and model settings. Then, we produce the best of our models via ablation studies. We finally compare our best model to other SOTA models in few-shot instance segmentation.

\subsection{Experiment Setup}

Our evaluation follows the works of few-shot object detection and instance segmentation \cite{tfa,imtfa}. We evaluate our approach on COCO \cite{coco}, VOC2007 \cite{everingham2010pascal}, and VOC2012 \cite{everingham2015pascal}. We divide 80 COCO classes into 60 base classes and 20 novel classes that intersect with classes in VOC. 5k images from COCO training and validation for testing while the remaining images are used for training. we conduct the experiments on datasets of having $K$ = 1, 5, 10 shots per novel class. We compare our appoarch (SMS + LR) with other methods in FSIS: Meta RCNN~\cite{meta-rcnn}, FGN~\cite{fgn}, MTFA~\cite{imtfa}. In addition, we also compare with SRR-FSD~\cite{zhu2021semantic} in FSOD. Our few-shot evaluation is based on~\cite{imtfa}.

\begin{figure*}[!t]
    \centering
    \includegraphics[width=\linewidth]{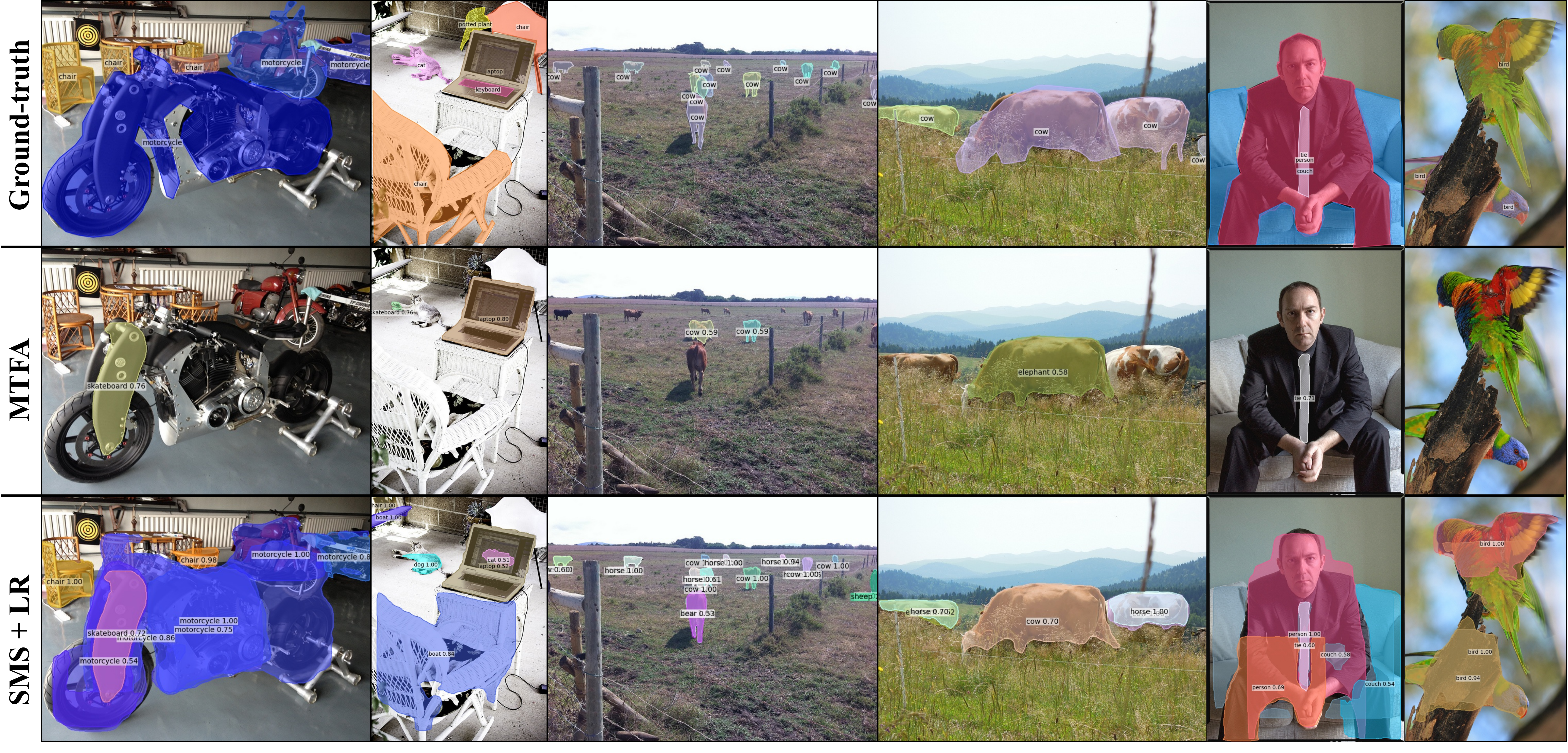}
    \caption{Visualization of the results on COCO of MTFA and our superclass approach under the 10-shot setting. Please view in color version with zoom.}
    \label{fig:visualization}
\end{figure*}

\subsection{Implementation Details}

We adopt Mask R-CNN \cite{he2017mask} with a combination of ResNet-101 \cite{he2016deep} and Feature Pyramid Network \cite{lin2017feature} as the backbone using Detectron2 \cite{wu2019detectron2}. During training, we apply standard data augmentation strategies including horizontal flipping, random cropping, and random resizing. Note that the methods in the data augmentation branch use their special data augmentation for few-shot learning. 

In the first base training phase, we use the SGD optimizer with an initial learning rate of 0.01, a mini-batch size of 8, momentum of 0.9, and the weight decay of 0.0001. The model is trained with 220,000 iterations on two GPUs and the learning rate is divided by 10 at 160,000 and 200,000 iterations. In the fine-tuning phase, the pre-trained model is frozen  except the last layers in prediction modules. To optimize the efficiency, we only fine-tune the parameters  with an initial learning rate of 0.01, a mini-batch size of 16 in 8000 iterations. We set $\alpha = 10^2$ and $\beta = 30$ in Equation \ref{eq:novel_loss} to train the model in novel sets. In the main experiment, we choose $\gamma = 15$ and $\epsilon = 0.5$  in Equations \ref{eq:smoothing} and \ref{eq:refinement}. In order to ease our ablation study, we reduce the number of iterations to $4,000$ for all experiments in the novel fine-tuning phase.

Regarding the baseline, we consider TFA \cite{tfa} with a mask head as similar as in MTFA \cite{imtfa}. We further fine-tune and report the MTFA* by using our backbone without class-agnostic, and its hyper-parameters are tuned to optimize the few-shot instance segmentation task.

\subsection{Ablation Study}
We conduct ablation studies on 10-shot of COCO dataset. We first show the performance of superclass variants in the Table \ref{table:ablations_framework_variants}. Then, we present the effectiveness of the number of base anchors on the result.

In Table \ref{table:ablations_framework_variants}, we show  AP and AP50 of the novel classes for each variant mentioned in Section \ref{subsec:Framework Variants}. It can be seen that due to ``hard" decisions, HSS makes more incorrect predictions of objects in novel classes than the others. With the improved version of HSS as SSS, the metrics show that it is important to soften the predictions of superclass classifiers. SSS gains double performance of HSS in the novel domain (approximately 3.0\% AP improvement). On the other hand, the final variant, SMS, succesfully leverages the relevant information of multiple base anchors to create new features. And the fact is that with $\gamma=2$, the SMS model achieves better results than the previous version SSS (+3.5\% AP), which just uses a single base anchor on novel classes. In addition, thanks to this advantages of relevant information, the AP on base classes is also enhanced about 2-3\% in comparison with 2 remaining variants.

We observe that increasing $\gamma$ can reduce the generalization of the model over the new domain in Table \ref{table:ablations_framework_variants}. This problem can be solved with LR (explained in Section \ref{subsec:LR}). In Table \ref{tab:Ablations_gamma}, we provide results of changing $\gamma$ when LR is applied in AP and AP50 . When the value of $\gamma$ is 2, 3 or 5, the model segments objects better than ones without applying LR. In terms of AP, SMS with LR improves 0.6\%, 1.27\%, 3.58\% in segmentation with $\gamma = 2, 3$ or $5$, respectively. This situation also occurs for experiments in detection.

In particular, keeping the changes of $\gamma$ increasingly helps to increase performance on novel data and even when $\gamma$ = 3 or 5 in contrast to Table \ref{table:ablations_framework_variants}. It demonstrates that the combination of SMS and LR can learn better the related information in a group. We achieve the best results as $\gamma$ = 15 with 11.80\% AP, 21.06\% AP50 in segmentation and 11.74\% AP, 21.86\% AP50 in detection for ablation configuration. As a result, we choose $\gamma$ = 15 for our main experiments in the next section. 

To clarify  the impact of $\gamma$ in our model,  we further report the performance of SMS and SMS+LR while gradually increasing the value of $\gamma$ in Fig.  \ref{fig:SMS_SMS+LR} with 10-shot data.

In the initial steps, compared with MTFA, we can see that the superclass model achieves outstanding results in AP50 with/without LR (i.e., SMS+LR and SMS). Then, SMS+LR demonstrates its ability in novel classification by gaining a rise to about 22\%. Nevertheless, when $\gamma$ progressively gets higher values, SMS+LR has a slight descend and SMS starts to have a rapid drop in performance. The potential reason for this drop is the increasing confusion between fine-grained classes. Specifically, within the same superclass, the incorrect associations create the ambiguity between fine-grained classes, and their influence is increasing as the value of $\gamma$ is raised.

\begin{table*}[!t]
\begin{center}
\begin{tabular}{l|l|ccc|ccc}
\bottomrule
\multirow{2}{*}{Supercategory*} & \multirow{2}{*}{Class} & \multicolumn{3}{c|}{Segmentation} & \multicolumn{3}{c}{Detection} \\
                               &                        & MTFA     & SMS+LR   & Improve.   & MTFA    & SMS+LR  & Improve.  \\\hline
Person                         & Person                 & 3.22     & 6.11     & 2.88       & 4.28    & 7.79    & 3.51      \\\hline
\multirow{7}{*}{Vehicle}       
                               & Car                    & 40.65    & 44.65    & 4.00       & 41.61   & 45.48   & 3.86      \\
                               & Airplane               & 39.12    & 45.48    & 6.36       & 38.85   & 45.95   & 7.10      \\
                               & Bus                    & 27.50    & 49.67    & 22.17      & 27.85   & 50.27   & 22.42     \\
                               & Motor.                 & 23.03    & 22.27    & -0.75      & 27.73   & 28.51   & 0.77      \\
                               & Train                  & 19.92    & 36.10    & 16.18      & 17.82   & 34.43   & 16.61     \\
                               & Boat                   & 12.42    & 15.04    & 2.62       & 10.13   & 13.25   & 3.12     \\
                               & Bicycle                & 1.54     & 3.31     & 1.77       & 2.69    & 4.61    & 1.92      
                               \\\hline
\multirow{6}{*}{Animal}        
                               & Cat                    & 23.52    & 27.29    & 3.77       & 20.97   & 25.52   & 4.55      \\
                               & Dog                    & 18.53    & 20.25    & 1.72       & 18.38   & 20.11   & 1.73      \\
                                & Cow                    & 16.98    & 17.38    & 0.40       & 18.48   & 19.83   & 1.35  \\
                               & Horse                  & 10.70    & 14.36    & 3.66       & 13.76   & 19.59   & 5.83      \\
                               & Bird                   & 11.89    & 12.67    & 0.79       & 10.99   & 11.63   & 0.63      \\
                               & Sheep                  & 8.87     & 8.91     & 0.04       & 9.30    & 10.47   & 1.17      
                                  \\\hline
Kitchen                        & Bottle                 & 12.38    & 17.79    & 5.41       & 12.56   & 18.32   & 5.76      \\\hline
\multirow{4}{*}{Furniture}     & Couch                  & 13.12    & 17.21    & 4.10       & 14.54   & 20.31   & 5.77      \\
                                & Chair                  & 3.27     & 8.20     & 4.93       & 4.77    & 10.17   & 5.40      \\
                               & Plant                  & 2.68     & 4.21     & 1.52       & 2.45    & 3.86    & 1.41      \\
                               & Table                  & 1.73     & 2.45     & 0.72       & 2.45    & 5.61    & 3.16      \\\hline
                               Electronic & TV                     & 56.65    & 59.46    & 2.80       & 55.69   & 59.57   & 3.88      \\\hline
\multicolumn{2}{c|}{Mean}                                & 17.39    & 21.64    & 4.25       & 17.76   & 22.76   & 5.00
\\\toprule
\end{tabular}
\end{center}
\caption{Performance per novel class under the 10-shot setting . Supercategory* is a property in COCO dataset.}
\label{tab:ap_per_class}
\end{table*}

\begin{table*}[!t]
    \begin{subtable}{0.5\linewidth}
    \centering
    \begin{tabular}[]{c|c|cccccc}
    \bottomrule
    Shots & Method                           & AP       & AP50     & AP75    \\\hline
    \multirow{2}{*}{10}    & {SRR-FSD} & 11.30    & \textbf{23.00}    & 9.80    \\
                           & SMS+LR                   & \textbf{12.31} & 22.76 & \textbf{11.60}   \\\hline
    \multirow{2}{*}{30}    &{SRR-FSD} & 14.70    & \textbf{29.20}    & 13.50   \\
                           & SMS+LR  & \textbf{15.11} & 27.09 & \textbf{14.90}  \\\toprule
    \end{tabular}
    \caption{}
    \label{table:additional_results}
    \end{subtable}
    \begin{subtable}{0.5\linewidth}
    \centering
    \begin{tabular}{l|cccc}
    \bottomrule
    \multicolumn{1}{l|}{Method} & FGN   & MTFA  & iMTFA & SMS            \\\hline
    AP                         & \_  & 9.51  & 8.57  & \textbf{10.12} \\
    AP50                       & 16.20 & 19.28 & 16.32 & \textbf{19.51} \\\toprule
    \end{tabular}
    \caption{}
    \label{tab:cross_domain}
    \end{subtable}
    \caption{(a) shows additional results for comparison with SRR-FSD \cite{zhu2021semantic}. The best performance is marked in boldface. Note that SRR-FSD is only applied in detection and the method also utilizes external text data. (b) shows FSIS results on cross-dataset. The symbol ``\_" refers to unavailable results.}
\end{table*}

\subsection{Comparisons with Baselines}
\label{subsec:compare_baseline}
We perform a comparison in COCO dataset and  the cross-dataset on VOC from COCO dataset (COCO2VOC) in FSIS and then an in-depth analyse is presented in this section.

\noindent\textbf{COCO.} The FSIS performance obtained by the various methods over shots comparatively reported in Table \ref{table:novel_set_coco}. In general, our approach which is a combination of SMS and LR get significant performance in comparison with the previous methods. We improve about 3\% AP and 6\% AP50 for shots when SMS+LR is compared with the baseline MTFA, especially 10-shot data (+4\% AP and +7\% AP50). The results demonstrate that our approach can strongly improve the FSIS methods by effectively leveraging the relation between base and novel classes. Compared to Meta RCNN and a fully-converged Mask RCNN model fine-tuned on the novel phase (MRCN+ft-full), SMS+LR surpasses all metrics with a large margin for every shot settings. In the strict metric AP75, our method improves 7\%, and 9\% in 5, 10-shot data while compared to Meta RCNN. 

To eliminate the effect of the backbone and tuning hyper-parameters, we provide the results of the reproduced MTFA* {(our baseline)} in Table \ref{table:novel_set_coco}. SMS+LR also outperforms these results of MTFA* which can be considered as the upper bound of MTFA. In comparison with recent methods, iFS-RCNN~\cite{nguyen2022ifs} and {RefT~\cite{han2023reference}}, our method achieves competitive  results in detection and even outperforms iFS-RCNN~\cite{nguyen2022ifs} (i.e., +1.48\% AP, +0.52\%AP, +2.13\%AP for 1, 5, 10-shot respectively) in segmentation. Note that RefT~\cite{han2023reference} applies a dual-branch architecture and a transformer framework to achieve impressive results. However, one limitation of RefT~\cite{han2023reference} is that the model is not able to perform well in a one-shot setting. Meanwhile, our method has achieved better results in terms of very scarce available training data (i.e. 1-shot setting), which indicates our method could be combined with transformer architecture to enhance the effectiveness when data is limited in the future work.

Table~\ref{table:novel_set_coco} also indicates that our method not only improves on large objects, but also well segments small and medium objects. This may be because the ambiguity of hard objects has been reduced when we take advantage of the base anchors that have worked well on the data domain with many variations to represent novel representatives. 

\begin{figure*}[!t]
    \centering
    \includegraphics[width=\textwidth]{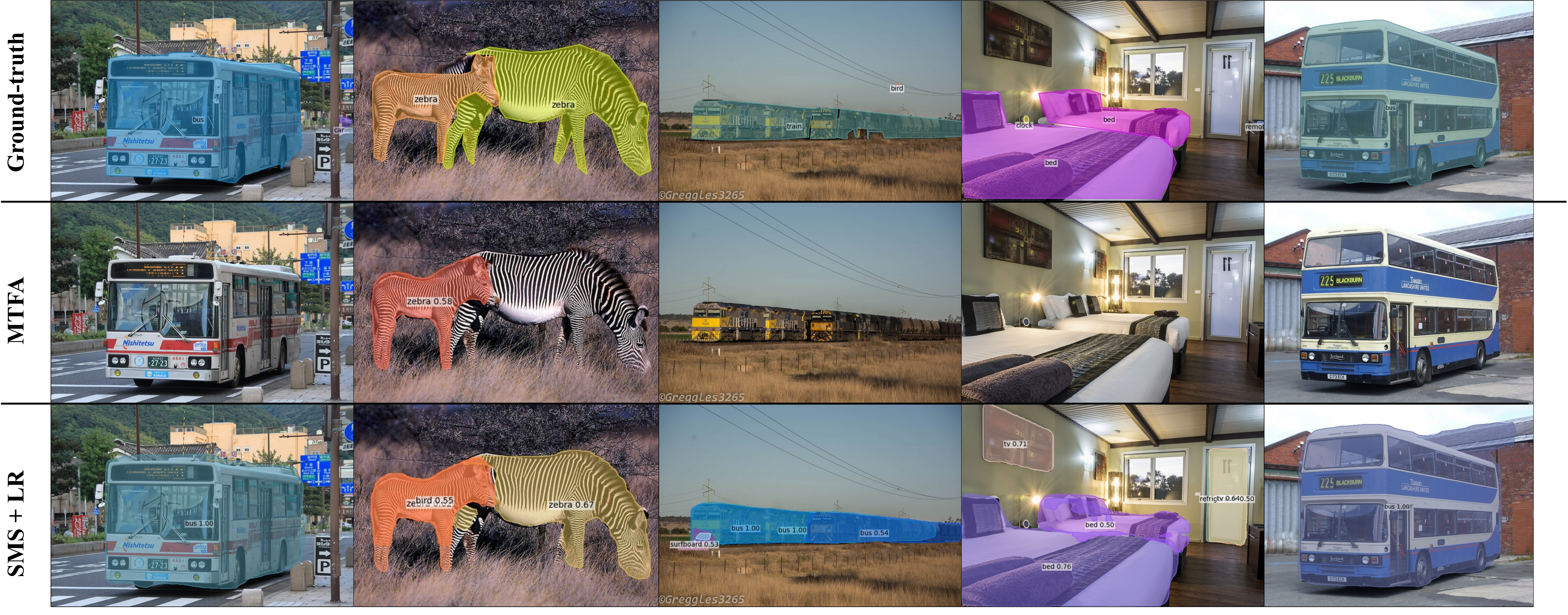}
    \caption{Visualization results of methods on 1-shot data.}
    \label{fig:sup_vis}
\end{figure*}

\begin{figure*}[!t]
    \begin{subfigure}{0.45\textwidth}
    \centering
    \includegraphics[width=\linewidth]{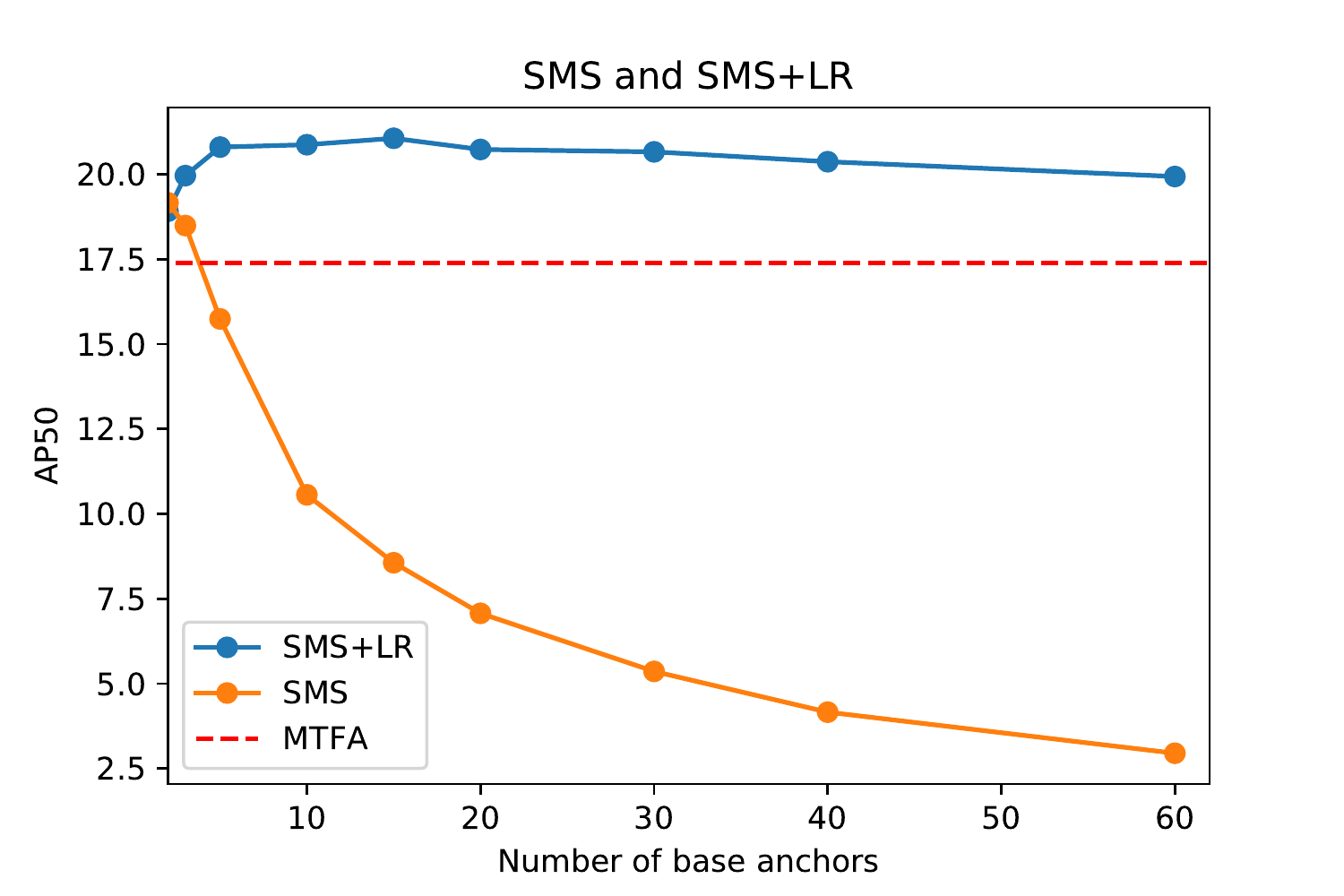}
    \caption{}
    \label{fig:SMS_SMS+LR}
    \end{subfigure}
    \centering
    \begin{subfigure}{0.45\textwidth}
    \includegraphics[width=\linewidth]{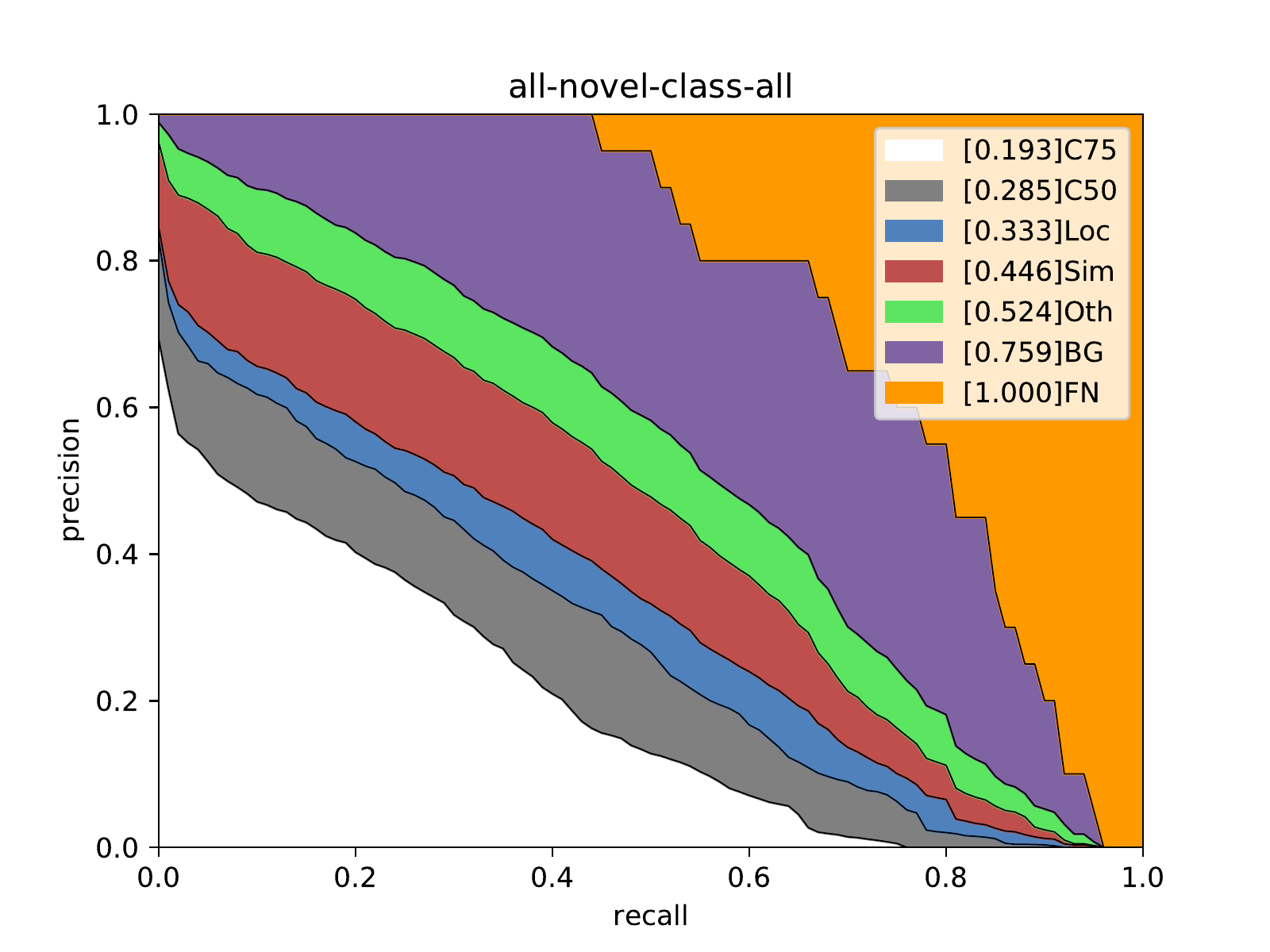}
    \caption{}
    \label{fig:pr-all}
    \end{subfigure}
    \caption{(a) performs AP50 for \textbf{novel classes} with different $\gamma$. (b) visualizes \textbf{precision-recall curve} for instance segmentation on the COCO novel classes.}
\end{figure*}

While data is very scarce as in 1 or 5-shot settings, our model also outperforms the previous method by about 3\% AP. This shows that the superclass approach can solve problems well with limited data. Indeed, as in  Table \ref{tab:fg_classes}, we see that the fine-grained classes for a superclass are highly similar over shots, which demonstrates the hierarchy to describe the association between classes for superclass models does not rely heavily on data. In addition, the superclass approach also creates groups that share many of the same features. For examples, the superclasses 1 represents features of vehicles, while the superclasses 3 and 4 contains the features of animal classes. Each superclass has the base anchor in the first position that is used to aid in describing the remaining fine-grained classes.

On the other hand, to demonstrate the efficiency of our model in the FSOD task, we also give a comparison with FSIS methods which also work on FSOD in Table \ref{table:novel_set_coco}. Via the performance, we also achieve superior results over other methods. Specifically, on the limited data like 1 shot, we obtain an improvement about 3\% for AP and 5\% for AP50. On more given data as 5 or 10 shots, the performance surpass the best of previous methods about 3-5\% and 4-7\%, respectively.

To better understand the effectiveness of our models, we also present the extra performance per class under the 10-shot setting. Table \ref{tab:ap_per_class} summarizes AP50 improvements in segmentation and detection. The results are sorted in decreasing order on the segmentation result. We follow the supercategories in COCO dataset to group classes. As shown in Table \ref{tab:ap_per_class}, classes in \texttt{\textbf{Vehicle}} (\texttt{Bus}, \texttt{Train}) or \texttt{\textbf{Electronic}} (\texttt{TV}) have a remarkable improvement. The reason behind this situation may be that the relevant characteristics of both \texttt{Truck} and \texttt{Bus} in \texttt{\textbf{Vehicle}} of COCO dataset are effectively explored by superclass models. The similar phenomenon can be observed with \texttt{TV} and \texttt{Laptop}.

In Table \ref{table:additional_results}, we further conduct the results for 30-shot data to compare with the SRR-FSD \cite{zhu2021semantic}, which is a recent work that also has the idea to exploit the similarity between classes for only FSOD. We even get competitive results SRR-FSD~\cite{zhu2021semantic}. Note that we only utilize the pretrained model instead of using external data as mentioned in~\cite{zhu2021semantic}. In 10-shot data, we achieve not only competitive results on AP50 but also higher than SRR-FSD about 1\% and 1.9\% on AP and AP75, respectively. The same situation goes for the 30-shot data.  These results indicate our approach outperforms ~\cite{zhu2021semantic} in exploiting the relation between classes. It can be due to the big gap in transferring information between different domains.

\begin{figure*}[!t]
    \begin{subfigure}{0.33\textwidth}
      \centering
    \includegraphics[width=\linewidth, trim={0.75cm 0 1.25cm 0}, clip]{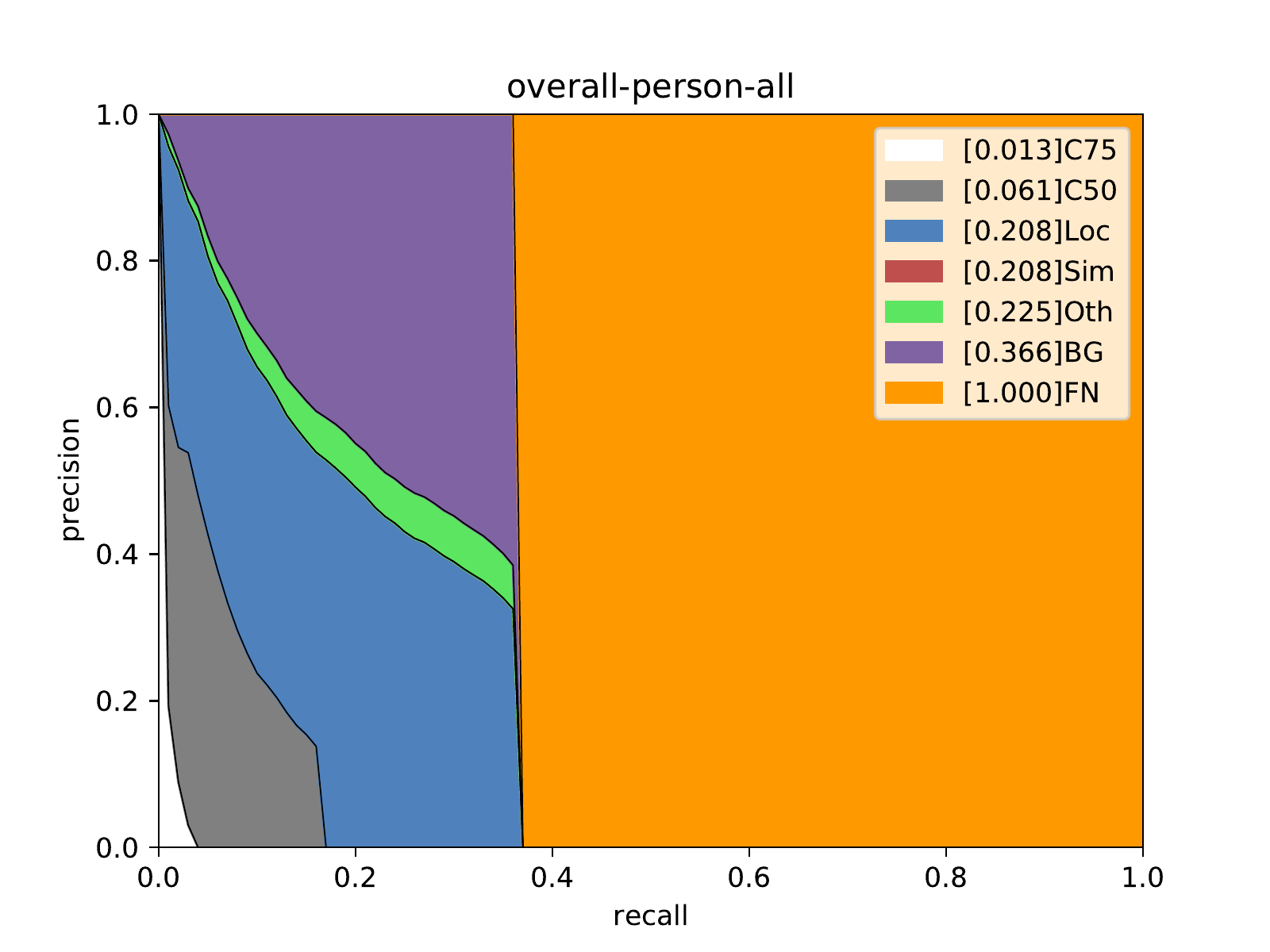}
    \caption{}
    \end{subfigure}
    \begin{subfigure}{0.33\textwidth}
      \centering
      \includegraphics[width=\linewidth, trim={0.75cm 0 1.25cm 0}, clip]{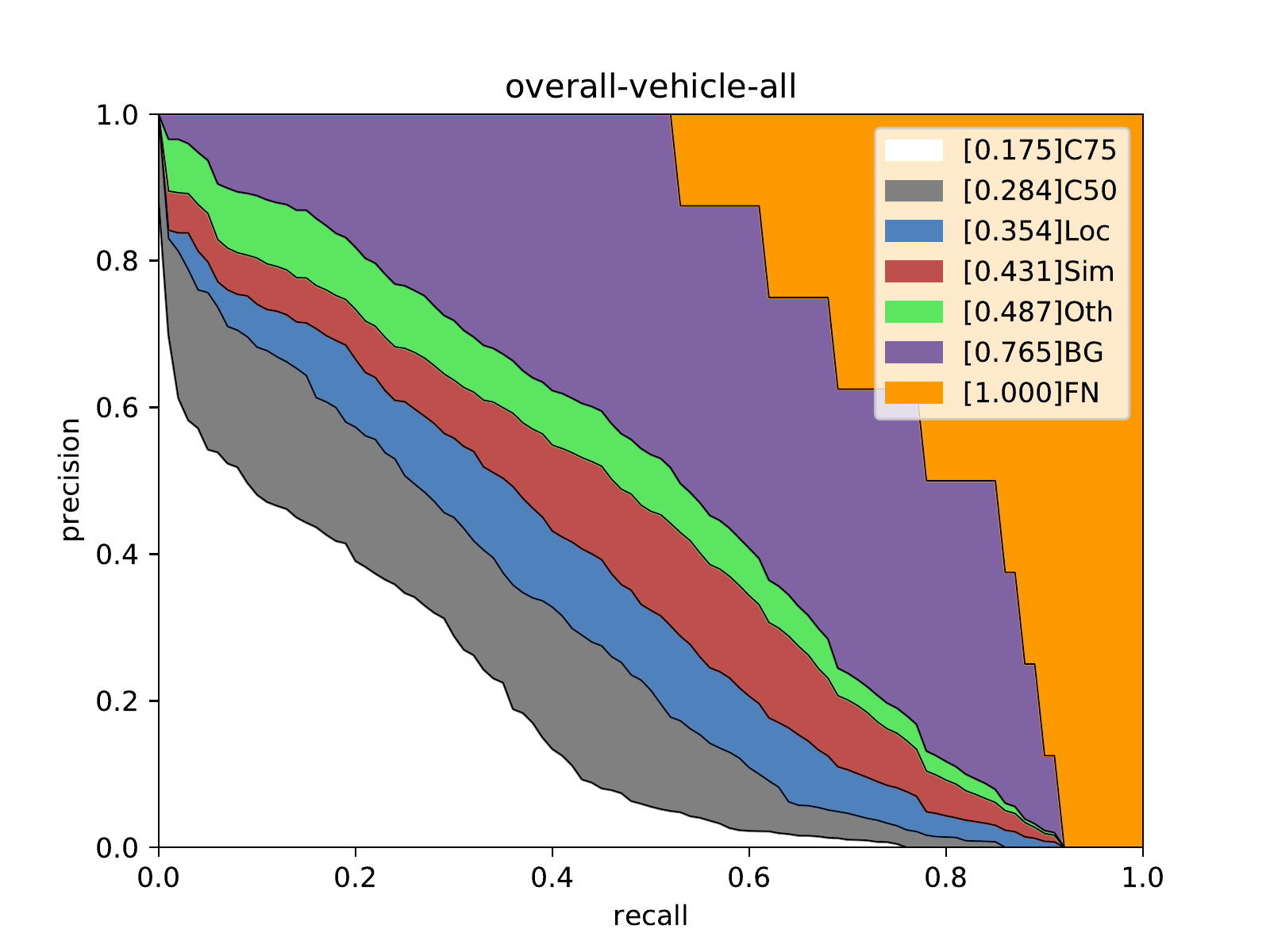}
      \caption{}
    \end{subfigure}
    \begin{subfigure}{0.33\textwidth}
      \centering
      \includegraphics[width=\linewidth, trim={0.75cm 0 1.25cm 0}, clip]{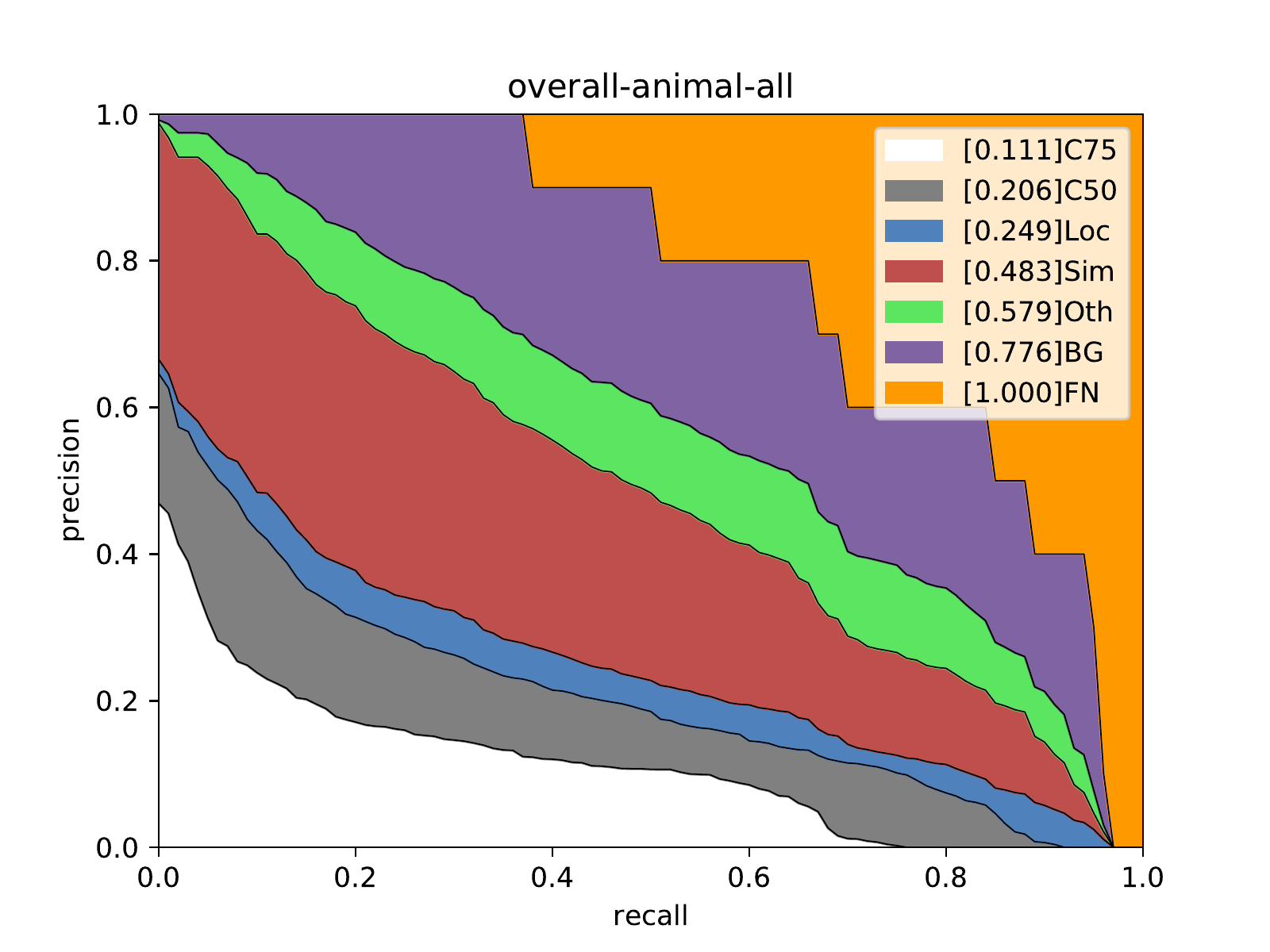}
      \caption{}
    \end{subfigure}   
    
    \begin{subfigure}{0.33\textwidth}
      \centering
    \includegraphics[width=\linewidth, trim={0.75cm 0 1.25cm 0}, clip]{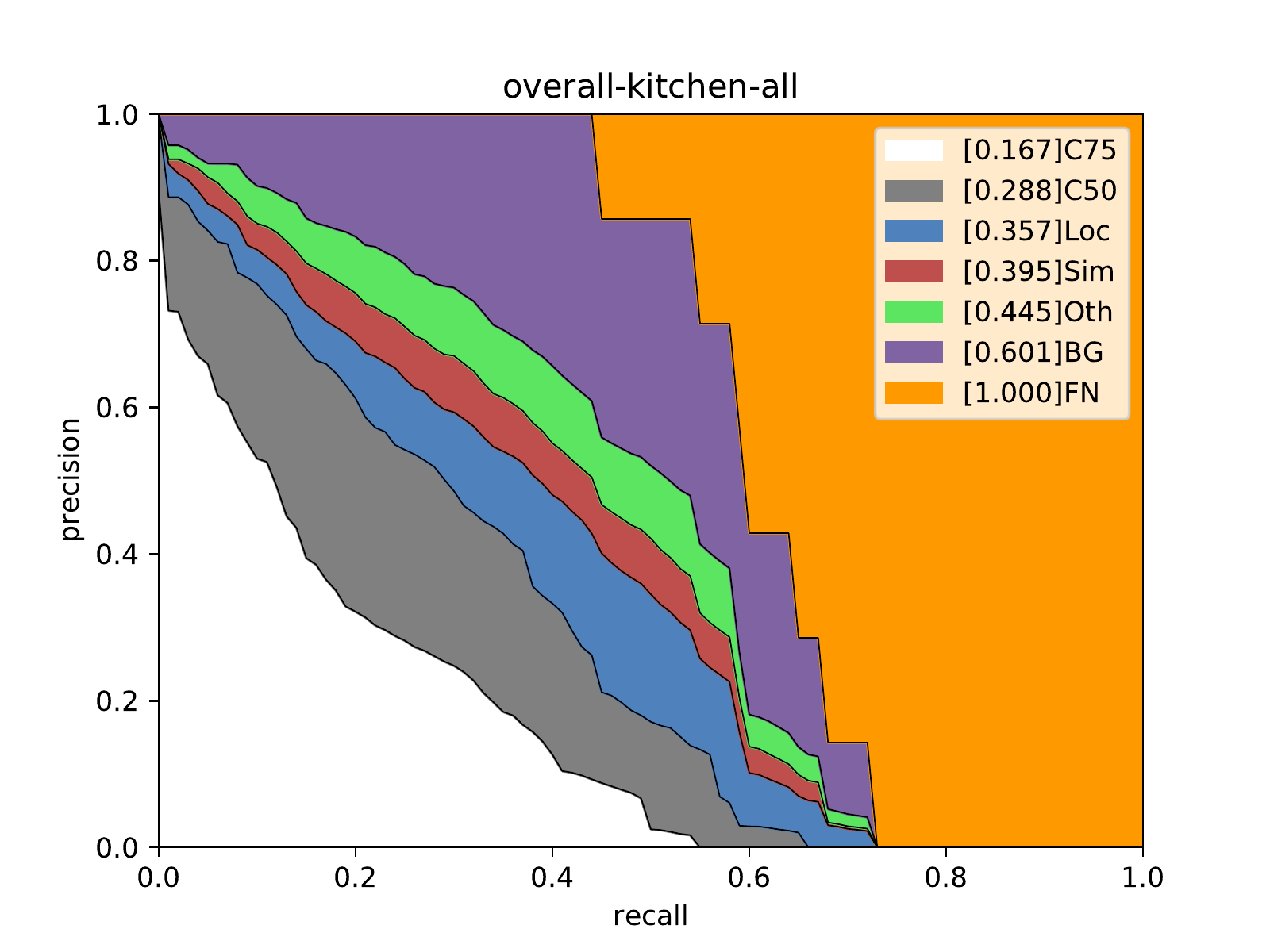}
    \caption{}
    \end{subfigure}
    \begin{subfigure}{0.33\textwidth}
      \centering
    \includegraphics[width=\linewidth, trim={0.75cm 0 1.25cm 0}, clip]{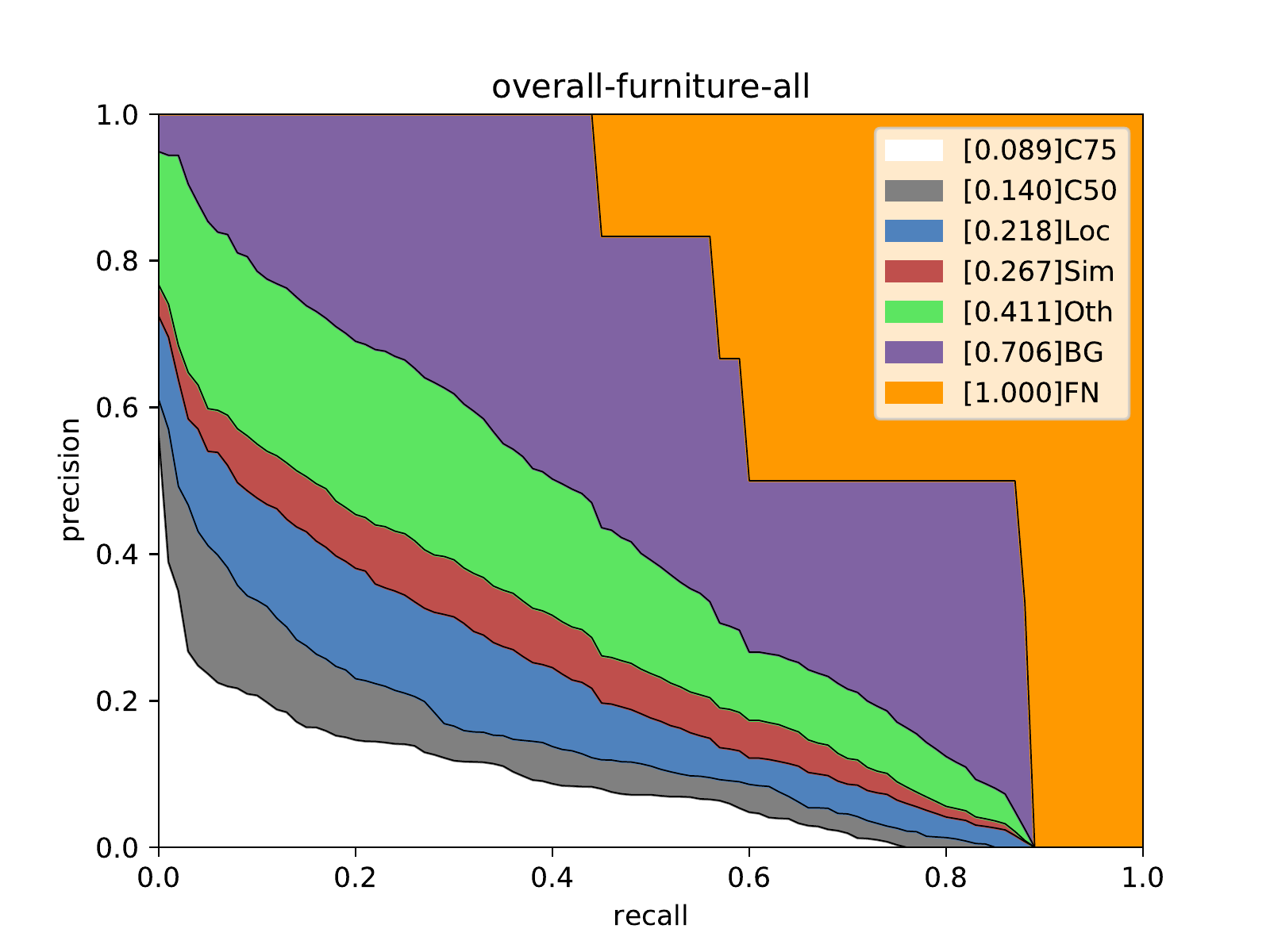}
    \caption{}
    \end{subfigure}
    \begin{subfigure}{0.33\textwidth}
      \centering
      \includegraphics[width=\linewidth, trim={0.75cm 0 1.25cm 0}, clip]{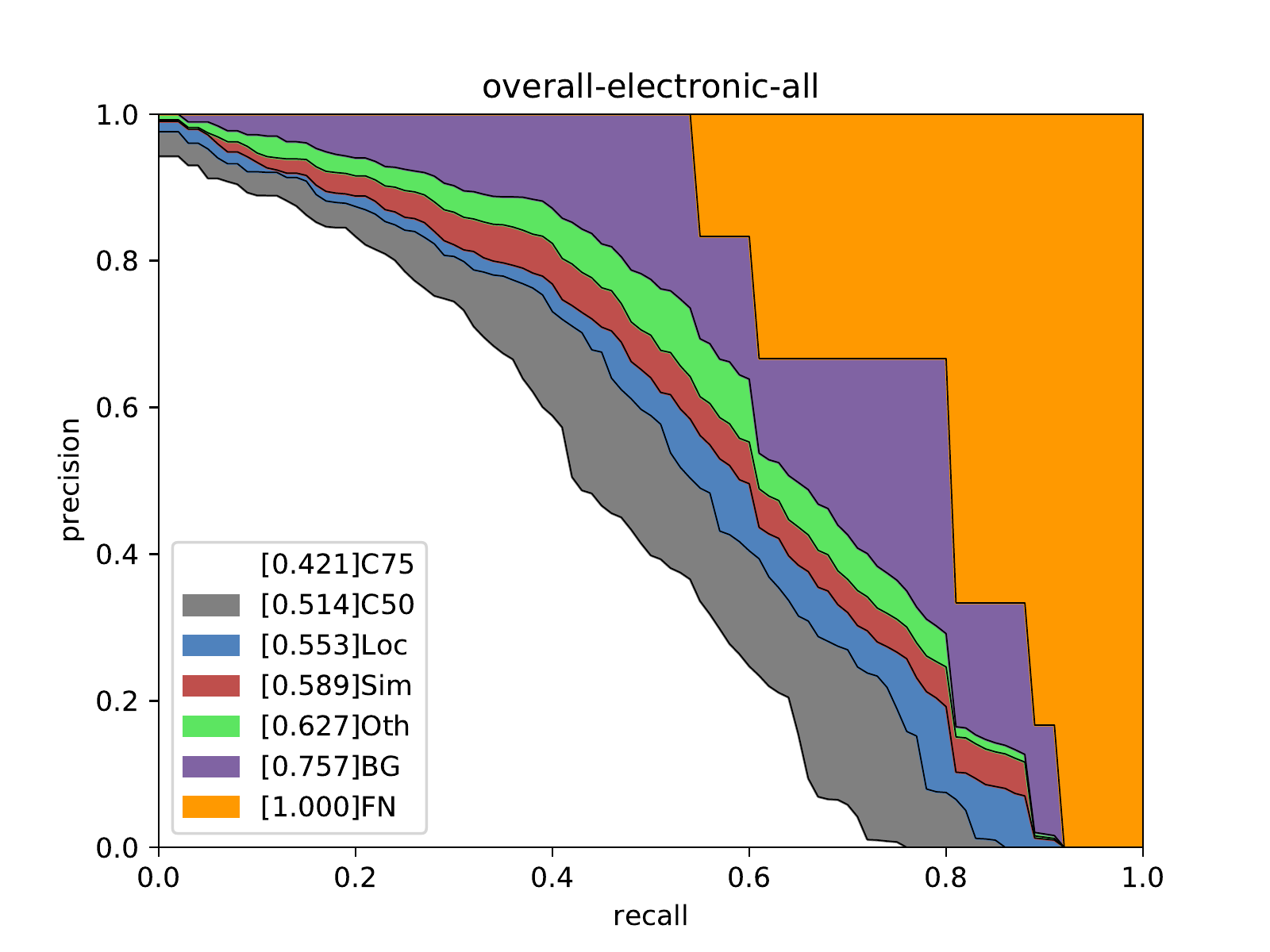}
      \caption{}
    \end{subfigure}

\caption{Precision-recall curves for instance segmentation on the supercategories of the COCO novel classes.}
\label{fig:super_curve}
\vspace{-5pt}
\end{figure*}

In addition, we present the qualitative results in Fig.  \ref{fig:visualization}. The results of the MTFA and our method under 10-shot setting are filtered on the threshold as 0.5 when compared to the ground-truth in COCO dataset. Overall, segmentation outcomes of our method are better than MTFA specifically small objects (third column), novel objects (three-right columns). However, our results can be over-segmented among fine-grained classes within a superclass. The results also show that FSIS is still very challenging and has room for further researches. 

We extend the quantitative results in Fig.  \ref{fig:sup_vis} which shows the visualization of results from our approach, and MTFA under 1-shot data. The figure demonstrates the superiority of our method when compared with MTFA in the context of the limited data.  In the first and the last columns, our approach has better segmentation for buses than MTFA. Still, our method gets over-segmentation in three mid-columns. Specifically, \texttt{Zebra} can be confused with \texttt{Bird}, or a pair of \texttt{Train} and \texttt{Bus}  when they are in the same supercategory. In the fourth column, our method classifies the thing on the wall as a TV and the door as a refrigerator because they have similarities in visual aspects.  The confident score results of MTFA are very low. Therefore, when we filter the result for visualization by a common threshold of 0.5, their results are removed.

\noindent\textbf{COCO2VOC.} Like FGN \cite{fgn} and MTFA \cite{imtfa}, we provide more results to demonstrate the generalization of our approach. In particular, we conduct the experiments on the cross-dataset VOC under \textit{1-way 1-shot} setting in Table \ref{tab:cross_domain}. The experiment indicates that our approach has superior performance in FSIS while compared to the previous methods.

\noindent\textbf{Class error analysis.} We also show the overall analysis~\cite{hoiem2012diagnosing} for novel classes in Fig.  \ref{fig:pr-all} and supercategories in Fig.  \ref{fig:super_curve}. The results in two mentioned Figures are gathered from SMS+LR under 10-shot data. Each visualization depicts a series of precision-recall curves where each PR curve is higher or equal to the previous ones. There are seven evaluation settings. The \textit{C75} and \textit{C50} curves represent the performance of methods at AP75 and AP50 metrics. The \textit{Loc} represents the precision of false positives which is caused by poor localization. The \textit{Sim} represents the confusion with similar objects in COCO supercategories, while the \textit{Oth} is confusion with other objects. The \textit{BG} is PR after all background false positives are removed. Finally, the \textit{FN} is PR after all remaining errors are removed. 

\noindent\textbf{Discussion.} We are aware that different branches of approaches such as data augmentation \cite{trans-int,khandelwal2021unit,max-margin}, or attention mechanism \cite{hu2021dense,rpn-attention} can achieve better results than our approach in few-shot object detection. As playing with legos, the above approaches can be applied in our framework to improve performance. Here, we provide some preliminary results of our method with spatial attention. In particular, we integrate an attention module (ASPP in \cite{chen2017rethinking}) into our model. The combination achieves 12.49\% AP, 24.23\% AP50 in detection, and 12.15\% AP, 22.51\% AP50 in segmentation while using 10-shot data for training. This indicates that our model is very flexible and it is possible that there will be improved versions by combining our method with others.

{\noindent\textbf{Limitations.} This work automatically builds a hierarchy relying on the similarity between base and novel class representations. Therefore, if the discrepancies between them are remarkable, features of base and novel classes in a superclass are diverse and high-variant. Hence, the representatives of the superclass do not well-represent fine-grained classes and assist the model to classify novel objects.

\vspace{-5pt}
\section{Conclusion and Future work}
\label{sec:conclusion}
 In this paper, we propose a new framework to address the problem of instance-level few-shot learning. In particular, our framework mines relevant information among given base classes in the class hierarchy  to describe novel objects via the proposed superclass. In addition, we apply the label refinement on fine-grained classes to smoothly describe the relationship between the base anchor and fine-grained classes and further improve the performance. The extensive experiments demonstrate the superiority of our proposed framework over the SOTA methods on the existing benchmarks. In the future,  we aim to reduce the phenomena of over-segmentation via post-processing methods and improve the model in the way that hierarchical information is efficiently mined.
 
 \section{Acknowledgement}
 This research was supported by the VNUHCM-University of Information Technology's Scientific Research Support Fund.

\balance
{
\bibliographystyle{splncs04}
\bibliography{egbib}

\begin{thebibliography}{10}
\providecommand{\url}[1]{\texttt{#1}}
\providecommand{\urlprefix}{URL }
\providecommand{\doi}[1]{https://doi.org/#1}

\bibitem{baek2021exploiting}
Baek, D., Oh, Y., Ham, B.: Exploiting a joint embedding space for generalized
  zero-shot semantic segmentation. In: Proceedings of the IEEE/CVF
  International Conference on Computer Vision (ICCV). pp. 9536--9545 (2021)

\bibitem{ben2021semantic}
Ben-Cohen, A., Zamir, N., Ben-Baruch, E., Friedman, I., Zelnik-Manor, L.:
  Semantic diversity learning for zero-shot multi-label classification. In:
  Proceedings of the IEEE/CVF International Conference on Computer Vision
  (ICCV). pp. 640--650 (2021)

\bibitem{cao2020few}
Cao, K., Ji, J., Cao, Z., Chang, C.Y., Niebles, J.C.: Few-shot video
  classification via temporal alignment. In: Proceedings of the IEEE/CVF
  Conference on Computer Vision and Pattern Recognition (CVPR). pp.
  10618--10627 (2020)

\bibitem{chen2017rethinking}
Chen, L.C., Papandreou, G., Schroff, F., Adam, H.: Rethinking atrous
  convolution for semantic image segmentation. arXiv preprint arXiv:1706.05587
  (2017)

\bibitem{chen2021elaborative}
Chen, S., Huang, D.: Elaborative rehearsal for zero-shot action recognition.
  In: Proceedings of the IEEE/CVF International Conference on Computer Vision
  (ICCV). pp. 13638--13647 (2021)

\bibitem{Cheng_2021_ICCV}
Cheng, J., Nandi, S., Natarajan, P., Abd-Almageed, W.: Sign:
  Spatial-information incorporated generative network for generalized zero-shot
  semantic segmentation. In: Proceedings of the IEEE/CVF International
  Conference on Computer Vision (ICCV). pp. 9556--9566 (October 2021)

\bibitem{cheraghian2021synthesized}
Cheraghian, A., Rahman, S., Ramasinghe, S., Fang, P., Simon, C., Petersson, L.,
  Harandi, M.: Synthesized feature based few-shot class-incremental learning on
  a mixture of subspaces. In: Proceedings of the IEEE/CVF International
  Conference on Computer Vision (ICCV). pp. 8661--8670 (2021)

\bibitem{everingham2015pascal}
Everingham, M., Eslami, S.A., Van~Gool, L., Williams, C.K., Winn, J.,
  Zisserman, A.: The pascal visual object classes challenge: A retrospective.
  International Journal of Computer Vision  \textbf{111}(1) (2015)

\bibitem{everingham2010pascal}
Everingham, M., Van~Gool, L., Williams, C.K., Winn, J., Zisserman, A.: The
  pascal visual object classes (voc) challenge. International Journal of
  Computer Vision  \textbf{88}(2) (2010)

\bibitem{rpn-attention}
Fan, Q., Zhuo, W., Tang, C.K., Tai, Y.W.: Few-shot object detection with
  attention-rpn and multi-relation detector. In: Proceedings of the IEEE/CVF
  Conference on Computer Vision and Pattern Recognition (CVPR) (2020)

\bibitem{fgn}
Fan, Z., Yu, J.G., Liang, Z., Ou, J., Gao, C., Xia, G.S., Li, Y.: Fgn: Fully
  guided network for few-shot instance segmentation. In: Proceedings of the
  IEEE/CVF Conference on Computer Vision and Pattern Recognition (CVPR) (2020)

\bibitem{fei2021z}
Fei, N., Gao, Y., Lu, Z., Xiang, T.: Z-score normalization, hubness, and
  few-shot learning. In: Proceedings of the IEEE/CVF International Conference
  on Computer Vision (ICCV). pp. 142--151 (2021)

\bibitem{imtfa}
Ganea, D.A., Boom, B., Poppe, R.: Incremental few-shot instance segmentation.
  In: Proceedings of the IEEE/CVF Conference on Computer Vision and Pattern
  Recognition (CVPR) (2021)

\bibitem{garg2022hiermatch}
Garg, A., Bagga, S., Singh, Y., Anand, S.: Hiermatch: Leveraging label
  hierarchies for improving semi-supervised learning. In: Proceedings of the
  IEEE/CVF Winter Conference on Applications of Computer Vision. pp. 1015--1024
  (2022)

\bibitem{gu2021lofgan}
Gu, Z., Li, W., Huo, J., Wang, L., Gao, Y.: Lofgan: Fusing local
  representations for few-shot image generation. In: Proceedings of the
  IEEE/CVF International Conference on Computer Vision (ICCV). pp. 8463--8471
  (2021)

\bibitem{han2023reference}
Han, Y., Zhang, J., Xue, Z., Xu, C., Shen, X., Wang, Y., Wang, C., Liu, Y., Li,
  X.: Reference twice: A simple and unified baseline for few-shot instance
  segmentation. arXiv preprint arXiv:2301.01156  (2023)

\bibitem{he2017mask}
He, K., Gkioxari, G., Doll{\'{a}}r, P., Girshick, R.B.: Mask {R-CNN}. In:
  Proceedings of the IEEE/CVF International Conference on Computer Vision
  (ICCV). pp. 2980--2988 (2017)

\bibitem{he2016deep}
He, K., Zhang, X., Ren, S., Sun, J.: Deep residual learning for image
  recognition. In: Proceedings of the IEEE/CVF Conference on Computer Vision
  and Pattern Recognition (CVPR) (2016)

\bibitem{hinz2022charactergan}
Hinz, T., Fisher, M., Wang, O., Shechtman, E., Wermter, S.: Charactergan:
  Few-shot keypoint character animation and reposing. In: Proceedings of the
  IEEE/CVF Winter Conference on Applications of Computer Vision (WACV). pp.
  1988--1997 (2022)

\bibitem{hoiem2012diagnosing}
Hoiem, D., Chodpathumwan, Y., Dai, Q.: Diagnosing error in object detectors.
  In: European conference on computer vision (ECCV). pp. 340--353. Springer
  (2012)

\bibitem{hong2021video}
Hong, J., Fisher, M., Gharbi, M., Fatahalian, K.: Video pose distillation for
  few-shot, fine-grained sports action recognition. In: Proceedings of the
  IEEE/CVF International Conference on Computer Vision (ICCV). pp. 9254--9263
  (2021)

\bibitem{hu2021dense}
Hu, H., Bai, S., Li, A., Cui, J., Wang, L.: Dense relation distillation with
  context-aware aggregation for few-shot object detection. In: Proceedings of
  the IEEE/CVF Conference on Computer Vision and Pattern Recognition (CVPR)
  (2021)

\bibitem{jhoo2021collaborative}
Jhoo, W.Y., Heo, J.P.: Collaborative learning with disentangled features for
  zero-shot domain adaptation. In: Proceedings of the IEEE/CVF International
  Conference on Computer Vision (ICCV). pp. 8896--8905 (2021)

\bibitem{yolo-reweighting}
Kang, B., Liu, Z., Wang, X., Yu, F., Feng, J., Darrell, T.: Few-shot object
  detection via feature reweighting. In: Proceedings of the IEEE/CVF
  International Conference on Computer Vision(ICCV) (2019)

\bibitem{khandelwal2021unit}
Khandelwal, S., Goyal, R., Sigal, L.: Unit: Unified knowledge transfer for
  any-shot object detection and segmentation. In: Proceedings of the IEEE/CVF
  Conference on Computer Vision and Pattern Recognition (CVPR) (2021)

\bibitem{kumar2019protogan}
Kumar~Dwivedi, S., Gupta, V., Mitra, R., Ahmed, S., Jain, A.: Protogan: Towards
  few shot learning for action recognition. In: Proceedings of the IEEE/CVF
  International Conference on Computer Vision (ICCV) Workshops. pp.~0--0 (2019)

\bibitem{lang2022learning}
Lang, C., Cheng, G., Tu, B., Han, J.: Learning what not to segment: A new
  perspective on few-shot segmentation. In: Proceedings of the IEEE/CVF
  Conference on Computer Vision and Pattern Recognition. pp. 8057--8067 (2022)

\bibitem{lee2022few}
Lee, H., Lee, M., Kwak, N.: Few-shot object detection by attending to
  per-sample-prototype. In: Proceedings of the IEEE/CVF Winter Conference on
  Applications of Computer Vision (WACV). pp. 2445--2454 (2022)

\bibitem{lengyel2021zero}
Lengyel, A., Garg, S., Milford, M., van Gemert, J.C.: Zero-shot day-night
  domain adaptation with a physics prior. In: Proceedings of the IEEE/CVF
  International Conference on Computer Vision (ICCV). pp. 4399--4409 (2021)

\bibitem{trans-int}
Li, A., Li, Z.: Transformation invariant few-shot object detection. In:
  Proceedings of the IEEE/CVF Conference on Computer Vision and Pattern
  Recognition (CVPR) (2021)

\bibitem{max-margin}
Li, B., Yang, B., Liu, C., Liu, F., Ji, R., Ye, Q.: Beyond max-margin: Class
  margin equilibrium for few-shot object detection. In: Proceedings of the
  IEEE/CVF Conference on Computer Vision and Pattern Recognition (CVPR) (2021)

\bibitem{li2020context}
Li, Y., Shao, Y., Wang, D.: Context-guided super-class inference for zero-shot
  detection. In: Proceedings of the IEEE/CVF Conference on Computer Vision and
  Pattern Recognition Workshops. pp. 944--945 (2020)

\bibitem{lin2017feature}
Lin, T.Y., Doll{\'a}r, P., Girshick, R., He, K., Hariharan, B., Belongie, S.:
  Feature pyramid networks for object detection. In: Proceedings of the
  IEEE/CVF Conference on Computer Vision and Pattern Recognition (CVPR) (2017)

\bibitem{coco}
Lin, T.Y., Maire, M., Belongie, S., Hays, J., Perona, P., Ramanan, D.,
  Doll{\'a}r, P., Zitnick, C.L.: Microsoft coco: Common objects in context. In:
  European conference on computer vision (ECCV) (2014)

\bibitem{panet}
Liu, S., Qi, L., Qin, H., Shi, J., Jia, J.: Path aggregation network for
  instance segmentation. In: Proceedings of the IEEE/CVF Conference on Computer
  Vision and Pattern Recognition (CVPR) (2018)

\bibitem{lu2021simpler}
Lu, Z., He, S., Zhu, X., Zhang, L., Song, Y.Z., Xiang, T.: Simpler is better:
  Few-shot semantic segmentation with classifier weight transformer. In:
  Proceedings of the IEEE/CVF International Conference on Computer Vision
  (ICCV). pp. 8741--8750 (2021)

\bibitem{mall2021field}
Mall, U., Hariharan, B., Bala, K.: Field-guide-inspired zero-shot learning. In:
  Proceedings of the IEEE/CVF International Conference on Computer Vision
  (ICCV). pp. 9546--9555 (2021)

\bibitem{memmesheimer2022skeleton}
Memmesheimer, R., H{\"a}ring, S., Theisen, N., Paulus, D.: Skeleton-dml: Deep
  metric learning for skeleton-based one-shot action recognition. In:
  Proceedings of the IEEE/CVF Winter Conference on Applications of Computer
  Vision (WACV). pp. 3702--3710 (2022)

\bibitem{oneshot-instance}
Michaelis, C., Ustyuzhaninov, I., Bethge, M., Ecker, A.S.: One-shot instance
  segmentation. arXiv preprint arXiv:1811.11507  (2018)

\bibitem{Min_2021_ICCV}
Min, J., Kang, D., Cho, M.: Hypercorrelation squeeze for few-shot segmentation.
  In: Proceedings of the IEEE/CVF International Conference on Computer Vision
  (ICCV) (2021)

\bibitem{mishra2018generative}
Mishra, A., Verma, V.K., Reddy, M.S.K., Arulkumar, S., Rai, P., Mittal, A.: A
  generative approach to zero-shot and few-shot action recognition. In:
  Proceedings of the IEEE/CVF Winter Conference on Applications of Computer
  Vision (WACV). pp. 372--380. IEEE (2018)

\bibitem{nguyen2022ifs}
Nguyen, K., Todorovic, S.: ifs-rcnn: An incremental few-shot instance
  segmenter. In: Proceedings of the IEEE/CVF Conference on Computer Vision and
  Pattern Recognition. pp. 7010--7019 (2022)

\bibitem{patravali2021unsupervised}
Patravali, J., Mittal, G., Yu, Y., Li, F., Chen, M.: Unsupervised few-shot
  action recognition via action-appearance aligned meta-adaptation. In:
  Proceedings of the IEEE/CVF International Conference on Computer Vision
  (ICCV). pp. 8484--8494 (2021)

\bibitem{faster-rcnn}
Ren, S., He, K., Girshick, R., Sun, J.: Faster r-cnn: towards real-time object
  detection with region proposal networks. IEEE transactions on pattern
  analysis and machine intelligence  \textbf{39}(6),  1137--1149 (2016)

\bibitem{shaban2022few}
Shaban, A., Rahimi, A., Ajanthan, T., Boots, B., Hartley, R.: Few-shot
  weakly-supervised object detection via directional statistics. In:
  Proceedings of the IEEE/CVF Winter Conference on Applications of Computer
  Vision (WACV). pp. 3920--3929 (2022)

\bibitem{sheynin2021hierarchical}
Sheynin, S., Benaim, S., Wolf, L.: A hierarchical transformation-discriminating
  generative model for few shot anomaly detection. In: Proceedings of the
  IEEE/CVF International Conference on Computer Vision (ICCV). pp. 8495--8504
  (2021)

\bibitem{sun2021fsce}
Sun, B., Li, B., Cai, S., Yuan, Y., Zhang, C.: Fsce: Few-shot object detection
  via contrastive proposal encoding. In: Proceedings of the IEEE/CVF Conference
  on Computer Vision and Pattern Recognition. pp. 7352--7362 (2021)

\bibitem{szegedy2016rethinking}
Szegedy, C., Vanhoucke, V., Ioffe, S., Shlens, J., Wojna, Z.: Rethinking the
  inception architecture for computer vision. In: Proceedings of the IEEE/CVF
  Conference on Computer Vision and Pattern Recognition (CVPR) (2016)

\bibitem{tian2020prior}
Tian, Z., Zhao, H., Shu, M., Yang, Z., Li, R., Jia, J.: Prior guided feature
  enrichment network for few-shot segmentation. IEEE transactions on pattern
  analysis and machine intelligence  (2020)

\bibitem{wang2022dynamic}
Wang, H., Liu, J., Liu, Y., Maji, S., Sonke, J.J., Gavves, E.: Dynamic
  transformer for few-shot instance segmentation. In: Proceedings of the 30th
  ACM International Conference on Multimedia. pp. 2969--2977 (2022)

\bibitem{tfa}
Wang, X., Huang, T.E., Darrell, T., Gonzalez, J.E., Yu, F.: Frustratingly
  simple few-shot object detection. In: ICML (2020)

\bibitem{wu2019detectron2}
Wu, Y., Kirillov, A., Massa, F., Lo, W.Y., Girshick, R.: Detectron2.
  \url{https://github.com/facebookresearch/detectron2} (2019)

\bibitem{wu2021learning}
Wu, Z., Shi, X., Lin, G., Cai, J.: Learning meta-class memory for few-shot
  semantic segmentation. In: Proceedings of the IEEE/CVF International
  Conference on Computer Vision (ICCV) (2021)

\bibitem{xie2021few}
Xie, G.S., Xiong, H., Liu, J., Yao, Y., Shao, L.: Few-shot semantic
  segmentation with cyclic memory network. In: Proceedings of the IEEE/CVF
  International Conference on Computer Vision (ICCV) (2021)

\bibitem{meta-rcnn}
Yan, X., Chen, Z., Xu, A., Wang, X., Liang, X., Lin, L.: Meta r-cnn: Towards
  general solver for instance-level low-shot learning. In: Proceedings of the
  IEEE/CVF International Conference on Computer Vision (ICCV) (2019)

\bibitem{Yang_2021_ICCV}
Yang, L., Zhuo, W., Qi, L., Shi, Y., Gao, Y.: Mining latent classes for
  few-shot segmentation. In: Proceedings of the IEEE/CVF International
  Conference on Computer Vision (ICCV) (2021)

\bibitem{zhang2021meta}
Zhang, C., Ding, H., Lin, G., Li, R., Wang, C., Shen, C.: Meta navigator:
  Search for a good adaptation policy for few-shot learning. In: Proceedings of
  the IEEE/CVF International Conference on Computer Vision (ICCV). pp.
  9435--9444 (2021)

\bibitem{zhong2022pica}
Zhong, C., Wang, J., Feng, C., Zhang, Y., Sun, J., Yokota, Y.: Pica: Point-wise
  instance and centroid alignment based few-shot domain adaptive object
  detection with loose annotations. In: Proceedings of the IEEE/CVF Winter
  Conference on Applications of Computer Vision (WACV). pp. 2329--2338 (2022)

\bibitem{zhu2021semantic}
Zhu, C., Chen, F., Ahmed, U., Shen, Z., Savvides, M.: Semantic relation
  reasoning for shot-stable few-shot object detection. In: Proceedings of the
  IEEE/CVF Conference on Computer Vision and Pattern Recognition (CVPR) (2021)

\end{thebibliography}
}

\vfill

\end{document}